\documentclass[a4paper,10pt]{article}

\usepackage{mathptmx}
\usepackage{graphicx}
\usepackage{times}
\usepackage[cmex10]{amsmath}
\usepackage{amssymb}
\usepackage[noend]{algorithmic}
\usepackage{algorithm}
\usepackage{overpic}
\usepackage{subfigure}
\usepackage{rotating}
\usepackage{tikz}
\usepackage{nicefrac}
\usetikzlibrary{arrows}

 \pdfoutput=1

\usepackage[colorlinks=true,linkcolor=blue,anchorcolor=blue,citecolor=blue,urlcolor=blue]{hyperref}

\title{Combinatorial Gradient Fields for 2D Images \\with Empirically Convergent Separatrices}

\author{ Jan Reininghaus \thanks{Institute for Science and Technology, Austria. \texttt{jan.reininghaus@ist.ac.at} }
\and     David G\"unther \thanks{MPI for Informatics, Germany. \texttt{\{dguenther, weinkauf, hpseidel\}@mpi-inf.mpg.de } }
\and     Ingrid Hotz \thanks{ Zuse-Insitute Berlin, Germany. \texttt{hotz@zib.de} }
\and     Tino Weinkauf \footnotemark[2] 
\and      Hans-Peter Seidel \footnotemark[2] }

\begin{document}

\maketitle
\begin{abstract}

This paper proposes an efficient probabilistic method that computes combinatorial gradient fields for two dimensional image data.
In contrast to existing algorithms, this approach yields a geometric Morse-Smale complex that converges almost surely to its continuous counterpart when the image resolution is increased.
This approach is motivated using basic ideas from probability theory and builds upon an algorithm from discrete Morse theory with a strong mathematical foundation.
%This approach is theoretically motivated using fundamental theorems from probability theory and elements from discrete Morse theory.
%
While a formal proof is only hinted at, we do provide a thorough numerical evaluation of our method and compare it to established algorithms.
\end{abstract}

%\author{Jan Reininghaus, David G\"unther, Ingrid Hotz, Tino Weinkauf, Hans-Peter Seidel% <-this % stops a space
%\IEEEcompsocitemizethanks{\IEEEcompsocthanksitem Jan Reininghaus is with IST Austria. E-mail: jan.reininghaus@ist.ac.at
%\IEEEcompsocthanksitem David G\"unther, Tino Weinkauf and Hans-Peter Seidel are with MPI for Informatics. E-mail: \{dguenther, weinkauf, hpseidel\}@mpi-inf.mpg.de
%\IEEEcompsocthanksitem Ingrid Hotz is with Zuse Institute Berlin. E-mail: hotz@zib.de}
%\thanks{}}

%\begin{keywords}
%Morse-Smale complex, discrete Morse theory, mathematical morphology.
%\end{keywords}}

\renewcommand{\c}[1]{~\cite{#1}}
\newcommand{\define}[1]{\emph{\bfseries{#1}}}

\newcommand{\alg}[1]{Algorithm~\ref{#1}}
\newcommand{\fig}[1]{Figure~\ref{#1}}
\newcommand{\tab}[1]{Table~\ref{#1}}
\newcommand{\chp}[1]{Chapter~\ref{#1}}
\renewcommand{\line}[1]{Line~\ref{#1}}
\renewcommand{\sec}[1]{Section~\ref{#1}}

\newcommand{\set}[2]{\{#1 :\, #2 \}}

\renewcommand{\algorithmicrequire}{\textbf{Input:}}
\renewcommand{\algorithmicensure}{\textbf{Output:}}

%\maketitle

\section{Introduction}
\label{sec:introduction}
%\IEEEPARstart{C}{omputer}
Computer assisted analysis of two-dimensional image data has become an essential tool in scientific research and industrial applications. To deal with the growing amount of data, automated feature extraction methods are frequently employed in applications from, e.g., medical imaging, geosciences or computer vision. In particular, methods based on computational topology \cite{edelsbrunner2010} are gaining traction due to their ability to robustly extract relevant features of the data. 

In this paper, we propose a novel method that extracts the Morse-Smale complex (MS-complex)\c{smale:gradientDynamicalSystems} of a given two-dimensional image. The MS-complex consists of critical points and separatrices. In this setting, the critical points are the local minimum-, saddle-, and maximum points, while the separatrices are the paths of steepest descent connecting the minima and maxima to the saddles\c{Cayley1859}. 

The MS-complex induces a segmentation of the image into regions of monotonic behavior\c{Milnor1965} and is strongly related\c{Najman1994} to the concept of the watershed transform\c{Maxwell1870}. In fact, the separatrices form a superset of the watersheds and watercourses\c{Griffin95,Lopez1999}.

There are three established methods to compute the MS-complex: The classical approach employs numerical methods. In this setting, the critical points are given by computing all zeros of the gradient. The separatrices are extracted by starting at the saddle points and following the gradient in the direction of the eigenvectors of their Hessian\c{Weinkauf2008a}. Using interval methods, the MS-complex can be extracted in a certified manner\c{Chattopadhyay2011}. The second approach works in a piecewise linear context. In this setting, the critical points are given by an analysis of the lower star of each vertex\c{Banchoff1970}. The separatrices are typically approximated as a sequence of steepest edges in the triangulation\c{zomordian:phdthesis, Bremer2004}.

In this paper, we build upon a purely combinatorial approach\c{Forman1998b,Forman2001} to compute the MS-complex. Such an approach lends itself to computational purposes due to its discrete nature\c{Bauer2011,Gyulassy2008a,Lewiner2005}. In contrast to the classical approach, it approximates the MS-complex directly on the grid defined by the image. In this setting, the critical points are defined by the topological changes in the sub-level sets of the data\c{milnor:morsetheory}. These topological changes can be computed efficiently by constructing a combinatorial gradient\c{Robins2011}. The separatrices are then computed by starting at the (combinatorial) saddle points and following the grid along this combinatorial gradient.

To analyze the approximation error of such a combinatorial algorithm, one can apply it to a sampling of an analytic function $f$ using $k^2$ pixels and compute the distance of its result to the exact MS-complex of $f$. A natural expectation is that this distance should go to zero when $k$ is increased. This is not the case for any established algorithm.

The main reason for this behavior is the combinatorial representation of the gradient direction in terms of the directions provided by the grid. In all established algorithms, the grid direction with the steepest descent is chosen to approximate the gradient. When one follows the combinatorial gradient, the quantization error can accumulate to a large value. Since this error only depends on the number of possible directions and the gradient of the data, it does not decrease when a finer grid is employed.

Based on the mathematical foundation presented in \sec{sec:background}, we propose an efficient probabilistic method that computes a combinatorial gradient in \sec{sec:method}. In contrast to all established algorithms, the approximation error of its induced MS-complex almost surely goes to zero when the image resolution is increased. While a formal proof is only hinted at, we do provide a thorough numerical evaluation of our method and compare it to established algorithms in \sec{sec:experiments}. We conclude this paper with a discussion of possible future directions and extensions in \sec{sec:conclusion}.

\section{Computational Discrete Morse Theory}
\label{sec:background}

This section introduces the main mathematical concepts and algorithms which our method builds upon. We will first give a brief introduction to discrete Morse theory\c{Forman1998b} in a graph theoretical notation\c{Reininghaus_TADD}. Using this notation, we will then describe the algorithm which we build upon in \sec{sec:method}.

\subsection{Definitions}
\label{sec:background_defintions}

Let $I$ denote a two dimensional image represented by real numbers defined on a rectangular grid $\Omega$. In topological terms, $\Omega$ is called a cubical cell complex $C$\c{Hatcher2002,Kaczynski2004}. This complex consists of cells with different dimensions (e.g., vertices, edges, cubes) and of boundary maps describing their neighborhood relation. For example, an edge is bounded by its two incident vertices, whereas a quad is bounded by its incident edges. 

To define the essential concepts of discrete Morse theory, we consider the cell complex $C$ in a graph theoretical setting:
the cell graph $G=(N,E)$ encodes the essential combinatorial information of $C$.
The nodes $N$ represent the cells of $C$
and each node $u^p$ is labeled by the dimension $p$ of the cell it represents.
The edges $E$ encode the neighborhood relation of the cells.
If a cell $u^p$ is in the boundary of a cell $w^{p+1}$,
then $e^p=\{u^p, w^{p+1}\} \in E$. The edge $e^p$ is said to be of index $p$.

The main task in computational discrete Morse theory is now to construct a combinatorial gradient such that the following combinatorial definitions of critical points and separatrices correspond to the input image $I$.

Formally, a combinatorial gradient field $V$ is a subset of pairwise non-adjacent edges of $G$ with a certain acyclic constraint\c{Chari2000} defined below. Given such a combinatorial gradient field $V$, the critical points are the unmatched nodes of $V$. A critical point $u^p$ that represents a cell of dimension $p$ is a minimum $(p=0)$, saddle $(p=1)$, or maximum $(p=2)$. 
A combinatorial $p$-line is a path in the cell graph $G$ whose edges are of index $p$ and alternate between $V$ and its complement $E\setminus V$. The above mentioned acyclic constraint is now specified as the non-existence of any closed $p$-line. A $p$-line connecting two critical points $u^p$ and $w^{p+1}$ is called a combinatorial $p$-separatrix. A $0$-separatrix thereby connects a minimum with a saddle, while a $1$-separatrix connects a saddle with a maximum.

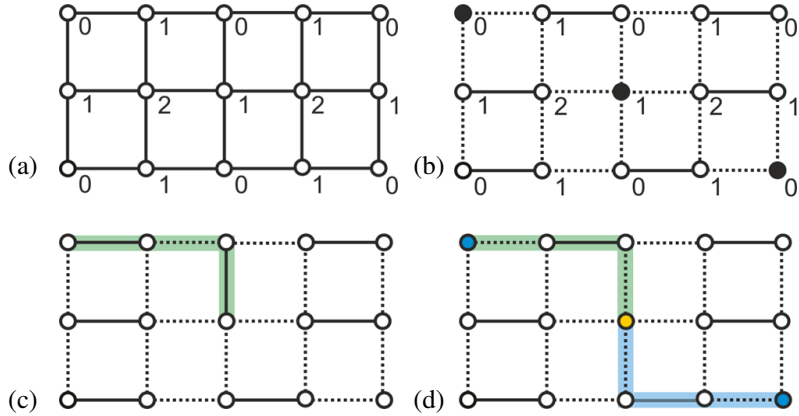
\begin{figure}%
\centering%
\begin{overpic}[width=0.8\linewidth]{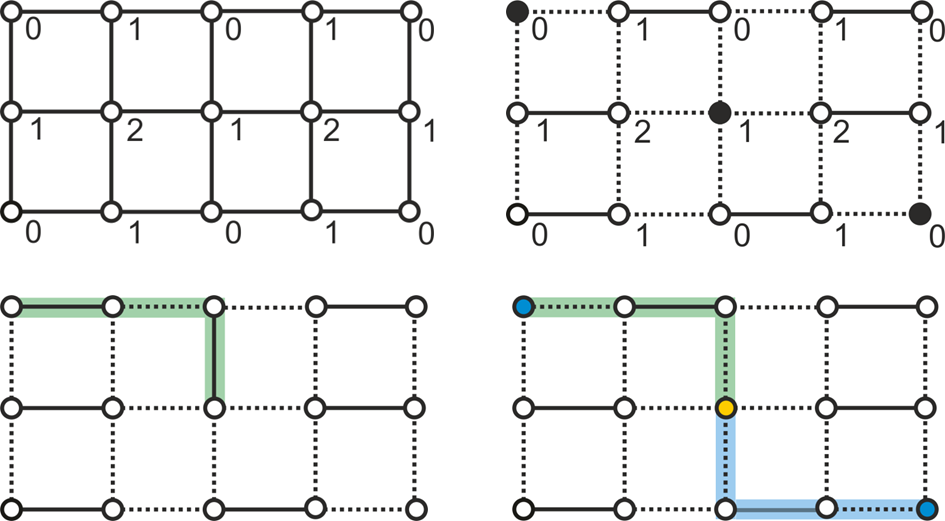}
\put(-7,32){(a)}
\put(48,32){(b)}
\put(-7,0){(c)}
\put(48,0){(d)}
\end{overpic}%
\caption{Illustration of a cell graph (a) of a $2\times1$ grid. The solid edges in (b) represent a combinatorial gradient containing critical points (black). A $p$-line is shown in (c), while (d) shows two separatrices (blue, green) connecting a saddle (yellow) to two minima (blue).}%
\label{fig:combinatorialGradient}%
\end{figure}

\fig{fig:combinatorialGradient} shows a simple cell graph of a $2\times 1$ grid and an arbitrary combinatorial gradient field with its critical points and separatrices. 

\subsection{Algorithm}

As already mentioned, the main task in computational discrete Morse theory is to construct a combinatorial gradient that corresponds to the input data $I$. Many such algorithms have been proposed~\cite{Bauer2010,gyulassi:valenz,King2005a,Lewiner2005}. In this paper, we make use of the algorithm \emph{ProcessLowerStars} proposed by Robins et al.~\cite{Robins2011}. The critical points of their combinatorial gradient provably correspond one-to-one to the topological changes of the lower-level sets of the input data in up to three dimensions. Also, this algorithm is very efficient, since it has linear running time and a parallel implementation scales well\c{Guenther_Topo3D}. 

In \sec{sec:method} we will propose an extension of this algorithm. Therefore, we now present it in detail in our graph theoretical notation.

We first propagate the input $I$ from the $0$-nodes to all nodes of the graph: each node $u^{p}$ is assigned the maximum $I$-value of the vertices of the cell that $u^{p}$ represents. We denote this extension of $I$ by $\hat{I}$. Algorithm~\ref{alg:CombinatorialGradient} decomposes then the cell graph $G$ into the lower stars~\cite{Banchoff1970} defined by $\hat{I}$ (line 3, 4, 5 and 6). Note that this decomposition is disjoint, which allows for good parallel scalability. Each lower star is now grown from its vertex using \emph{simple homotopic expansions} -- the inverse of homotopic collapses~\cite{cohen1973course}. Such expansions are represented by the edges of the cell graph (line~7).

The combinatorial gradient $V$ is now constructed iteratively (line 10): each time we expand a lower star using an edge $e \in E$ (lines 13, 14 and 15), we append $e$ to $V$ (line~16). An edge $e=\{u^{p},w^{p+1}\} \in E$ is admissible for simple homotopic expansion if the following conditions hold:
\begin{enumerate}
	\item $u^{p}$ and $w^{p+1}$ are not covered by an edge in the current $V$ (line 13),
	\item $u^{p}$ and $w^{p+1}$ have not been flagged previously (line 13),
	\item there is no other edge $\{z^{p},w^{p+1}\} \in E$ that fulfills 1. and 2. (line 14).
\end{enumerate}
If the set of admissible edges $L$ is empty (line 19), we flag an arbitrary node in the lower star (line 23) that is not covered by an edge in the current $V$ (line 22). If no such node can be found, the expansion stops. If $L$ is not empty (line 15), an admissible edge is chosen based on an order defined by $\hat{I}$ and appended to $V$ (line~16). 

As shown by Robins et al.~\cite{Robins2011}, the way we choose an edge from $L$ (line~\ref{SteepestDescent}) does not affect the overall number nor the type of critical points in the resulting combinatorial gradient. Since the combinatorial gradient is supposed to correspond to the (continuous) gradient, a natural choice is the edge that represents locally the steepest descent. 
\begin{algorithm}[ht]
  \caption{CombinatorialGradient$(G,I)$}
  \label{alg:CombinatorialGradient}
  \begin{algorithmic}[1]
  \REQUIRE $G=(N,E),\, \hat{I}: N \rightarrow \mathbb{R}$
  \ENSURE $V \subset E$
	\STATE $V \leftarrow \emptyset$
	\FORALL{ $v^0 \in N$}
		\STATE $S \leftarrow v^0$
		\STATE $W \leftarrow \set{w \in N}{\hat{I}(w) \leq \hat{I}(v^0)}$
		\FOR{ $p \leftarrow 0, \ldots, d-1$}
			\STATE $S \leftarrow S \cup \set{w^{p+1} \in W}{\exists \{u^{p},w^{p+1}\} \in E,\,u^{p} \in S}$
		\ENDFOR
		\STATE $K \leftarrow E(S)$
		\STATE $C \leftarrow \emptyset$
		\STATE $abort \leftarrow false$
		\WHILE{$abort = false$} \label{whileLoop}
			\STATE $abort \leftarrow true$
			\FOR{ $p \leftarrow 0, \ldots, d-1$}
				\STATE $T \leftarrow \set{\{u^{p},w^{p+1}\} \in K}{u^{p}, w^{p+1}\notin C \cup N(V)}$
				\STATE $L \leftarrow T \setminus \set{\{u^{p},w^{p+1}\}\in T}{\exists \{z^{p},w^{p+1}\} \in T,\, u^{p} \neq z^{p} } $
				\IF{ $L \neq \emptyset$ }
						\STATE $V \leftarrow V \cup \,\textcolor{red}{ChooseEdge(L)} $ \label{SteepestDescent}
						\STATE $abort \leftarrow false$
						\STATE \textbf{goto} \line{whileLoop}
				\ELSE
						\FOR{ $k \leftarrow 0, \ldots, d$}
							\STATE $\{u^{k}_0, u^{k}_1,\ldots, u^{k}_m\} \leftarrow	\set{u^{k}\in S}{u^{k} \notin C \cup N(V)}$
							\IF{ $\set{u^{k}\in S}{u^{k} \notin C \cup N(V)} \neq \emptyset$ }
								\STATE $C \leftarrow C \cup u^{k}_0$
								\STATE $abort \leftarrow false$
								\STATE \textbf{goto} \line{whileLoop}
							\ENDIF
						\ENDFOR
				\ENDIF
			\ENDFOR
		\ENDWHILE
	\ENDFOR
  \end{algorithmic}
\end{algorithm}

%As shown by Robins et al.\c{Robins2011}, the way we choose an edge from $L$ (\line{SteepestDescent} of \alg{alg:CombinatorialGradient}) does not affect the overall number and type of critical points in the resulting combinatorial gradient. It only affects the separatrices. Since the combinatorial gradient is supposed to correspond to the (continuous) gradient of the input, a natural choice is the edge that represents locally the steepest descent. However, this choice does not yield convergent separatrices in the sense of \sec{sec:introduction}. In the following section, we give an explanation for this behavior and present a strategy to choose edges from $L$ that yields convergent separatrices.
%
%

\section{Almost Surely Convergent Separatrices}
\label{sec:method}

Before we describe our method in \sec{sec:method_method}, we give an explanation for the non-convergent behavior of the existing combinatorial gradient algorithms and motivate our almost surely convergent probabilistic approach.

\subsection{Motivation}
\label{sec:method_motivation}

As defined in \sec{sec:background_defintions}, the separatrices of a combinatorial gradient $V$ are alternating paths in the cell graph $G$ with respect to $V$. Intuitively, the edges in $V$ should therefore reflect the direction of the gradient of the input data. However, at a given vertex $u$, there are only a constant number of directions representable by the edges of $G$. This implies that the continuous gradient can only be represented in a quantized way. Loosely speaking, the gradient direction is snapped to the edges of the graph. 

Therefore, the combinatorial gradient differs from the continuous gradient not only by a sampling error, but also by a quantization error. Note that the sampling error can be decreased using a denser sampling. However, this is not necessarily the case for the quantization error. 

Perhaps surprisingly, the ubiquitous steepest descent strategy for the edge selection in \line{SteepestDescent} of \alg{alg:CombinatorialGradient} suffers from this quantization artifact. 
Suppose that the exact gradient is almost constant in a region $K$ and points 'North-North-East'. At any given vertex in $K$ the steepest descent direction is therefore always 'North' -- independent of the resolution used to sample the exact gradient. Any exact separatrix passing through $K$ is thereby approximated by a straight line going 'North'. This (resolution independent) behavior can be observed frequently in practice as can be seen in \fig{fig:randomFields}. 

To deal with the quantization error, we propose to choose the edges in \line{SteepestDescent} of \alg{alg:CombinatorialGradient} adjacent to a vertex $u$ in a probabilistic fashion. Since we cannot represent the (continuous) gradient direction exactly, we pick an edge according to a random variable $X_u$. The probability mass function $g$ of $X_u$ is defined by the image data $I$ and the width and height of each pixel. The basic idea is to design $g$ such that the expected value of $X_u$ corresponds to the (continuous) gradient direction at $u$. 

Note that these random variables are independent. Assuming that in this setting the law of large numbers\c{Bernoulli} is applicable, a path following this probabilistic combinatorial gradient will therefore almost surely proceed in the direction of the (continuous) gradient when the grid is refined. While this argument is far from a formal proof, we do provide a thorough numerical evaluation in \sec{sec:experiments} that substantiates this intuition.

\subsection{Method}
\label{sec:method_method}

The main building block of our method consists of \alg{alg:CombinatorialGradient}. The only change is that we choose the direction (\line{SteepestDescent} of \alg{alg:CombinatorialGradient}) from $L$ in a probabilistic fashion instead of choosing the locally steepest descent. 

The index of the edges in $L$ is either always $1$ or always $0$ (Line 14 of \alg{alg:CombinatorialGradient}). Perhaps surprisingly, it suffices to choose the edges of index $0$ appropriately. The order in which the edges of index $1$ are chosen has no effect. They are uniquely defined once the edges of index $0$ and the saddle points have been selected. This fact motivated the original construction of a combinatorial gradient in\c{Lewiner2003}. In the following, we therefore assume that $L$ contains only edges of index $0$. 

 For a given vertex $u^0\in N$, the edge selection strategy is given by a random variable $X_u$. The value of this random variable is always an edge in $L$. 
We now define a probability mass function $g: L \rightarrow [0,1]$ for $X_u$ such that the expected value of $X_u$ is collinear to the (continuous) gradient at $u$.

We assume (without loss of generality) that the (continuous) gradient points North-East, i.e.,  $\nabla I (u) = (I_x,I_y)$ with $I_x, I_y \geq 0$. Furthermore, the width of the current pixel is denoted by $w$ and its height by $h$. The set $L$ thereby consists of the directions $(0,h)$ and $(w,0)$. 

To simplify notation, we refer to $g((w,0))$ by $\lambda$. Since $g$ is a probability mass function, we have $g((0,h)) = 1 - \lambda$. The expected direction $E(X_u)$ is now given by
\begin{equation}
E(X_u) =  (1-\lambda)\left( \begin{array}{c}  0 \\  h \end{array} \right) 
+ \lambda  \left( \begin{array}{c} w \\ 0 \end{array} \right)  
= \left( \begin{array}{c} \lambda w \\ (1-\lambda) h \end{array} \right).
\label{eq:expected_value}
\end{equation}

Since $E(X_u)$ should be collinear to $\nabla I (u) = (I_x,I_y)$, the following condition must hold
\begin{equation}
\det \left( 
\begin{array}{cc}
\lambda w & I_x \\
(1-\lambda) h & I_y
\end{array}
\right) = 0.
\label{eq:collinear_condition}
\end{equation}

This yields
\begin{equation}
\begin{aligned}
g\left((w,0)\right)  &= \frac{h I_x}{w I_y + h I_x} \text{, and } g\left((0,h)\right)  &= \frac{w I_y}{w I_y + h I_x}.
\label{eq:probabilities}
\end{aligned}
\end{equation}

Since $\nabla I (u)$ is not directly available, we approximate it using finite differences: $ $
\begin{equation}
\begin{aligned}
I_x \approx \frac{I((u + (w,0)) - I(u)}{w} \text{, and } I_y \approx \frac{I((u + (0,h)) - I(u)}{h}.
\label{eq:finite_differnences}
\end{aligned}
\end{equation}

Denoting the height difference $I((u + (w,0)) - I(u)$ by $W$ and $I((u + (0,h)) - I(u)$ by $H$, and inserting \eqref{eq:finite_differnences} into \eqref{eq:probabilities} yields the final probability mass function $g$ in terms of $I$, $w$ and $h$:
\begin{equation}
\begin{aligned}
g\left((w,0)\right)  &= \frac{h^2 W	}{w^2 H + h^2 W} \text{, and } g\left((0,h)\right)  &= \frac{w^2 H}{w^2 H + h^2 W}.
\label{eq:final_result}
\end{aligned}
\end{equation}

Note that, in practice, $L$ may consist of more than $2$ edges due to the sampling of $I$. Each edge in $L$ is therefore assigned the height difference weighted by the squared length of the dual edge. Its probability is then given by the normalized value with respect to the other edges in $L$.

\section{Evaluation and Comparison}
\label{sec:experiments}

In the following, we evaluate our probabilistic method and compare it to established algorithms. All experiments were performed on a machine with two Intel Xeon E5645 CPUs. We applied the linear \alg{alg:CombinatorialGradient} in parallel. For an image of resolution $4096^2$, \alg{alg:CombinatorialGradient} needed about 6 seconds. The probabilistic and the steepest descent version take the same time, since the amount of time needed to choose the edges is negligible. Using the implicit representation of the cell graph proposed in\c{Guenther_Topo3D}, the memory requirement is very low. To process an image of $k$ pixels we need $2k$ bytes of main memory. 

\begin{figure*}[!t]%
\centering%
\begin{overpic}[width=0.14\linewidth]{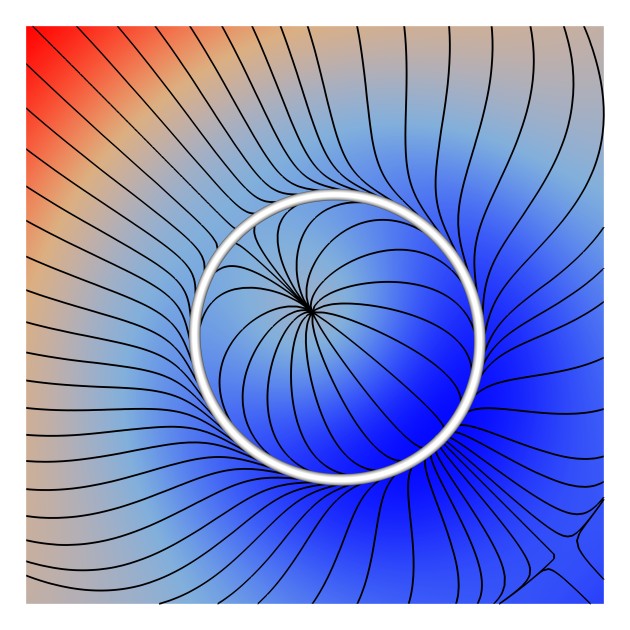} 
  \put(-5,30){\begin{rotate}{90}\tiny$f(x,y)$\normalsize\end{rotate}}
 \end{overpic}\;%
\begin{overpic}[width=0.14\linewidth]{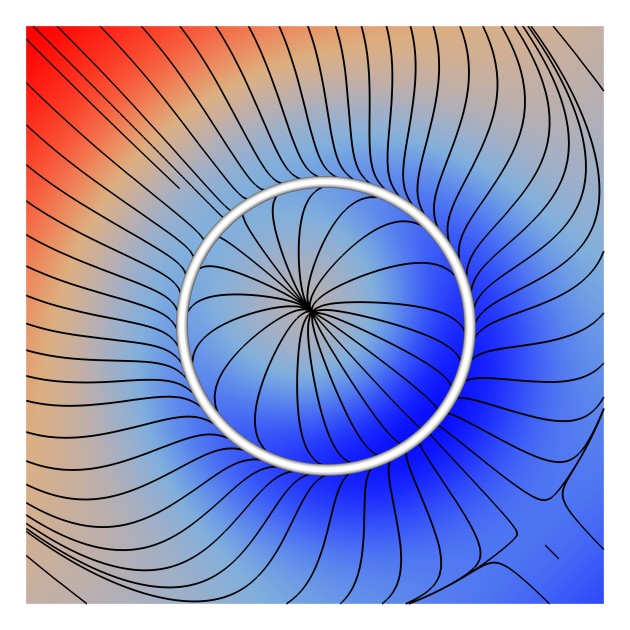} \end{overpic}\;%
\begin{overpic}[width=0.14\linewidth]{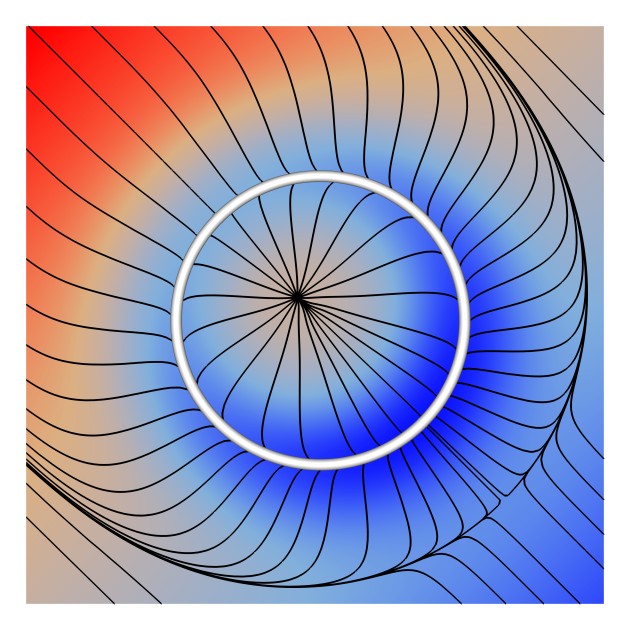} \end{overpic}\;%
\begin{overpic}[width=0.14\linewidth]{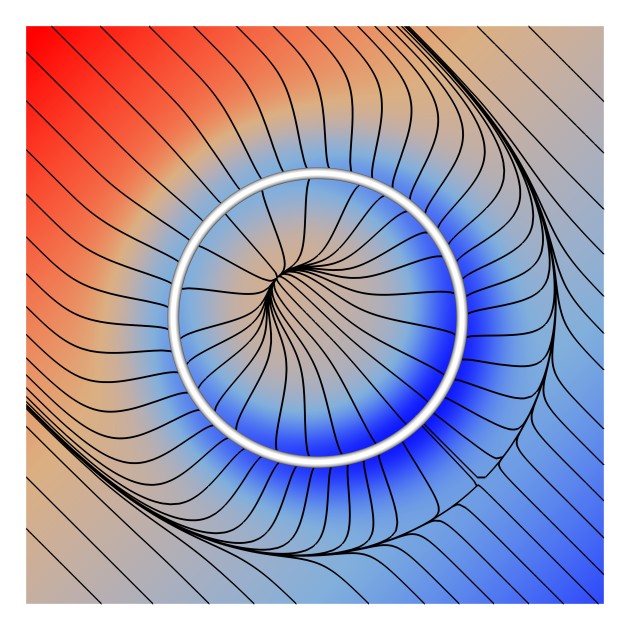} \end{overpic}\;%
\begin{overpic}[width=0.14\linewidth]{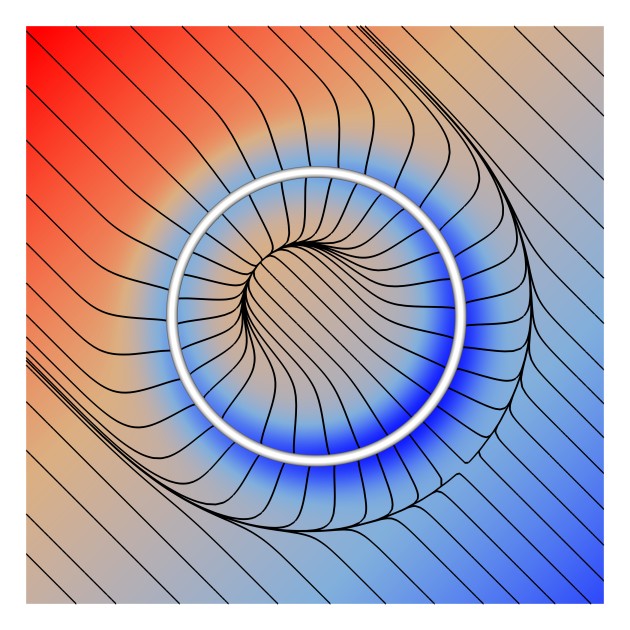} \end{overpic}\;%
\begin{overpic}[width=0.14\linewidth]{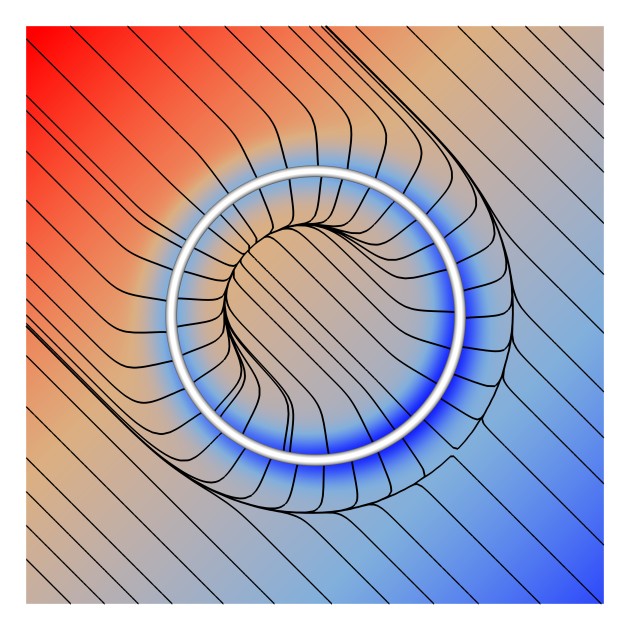} \end{overpic}\vspace{-0.5ex}\\%
\begin{overpic}[width=0.14\linewidth]{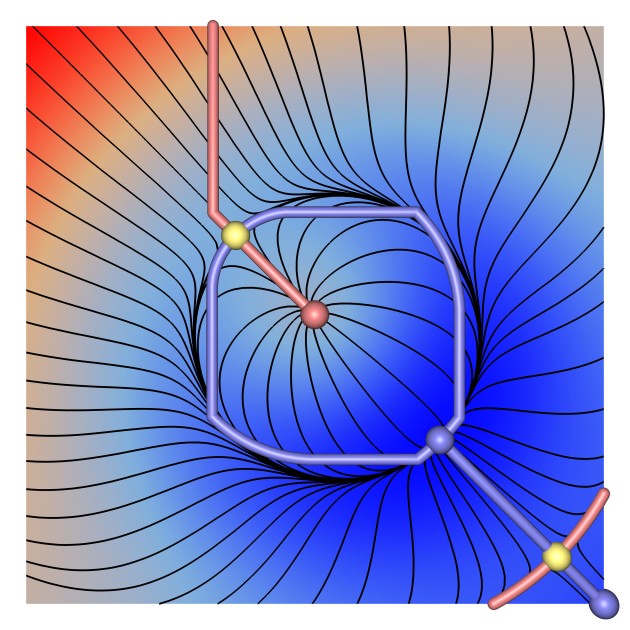} 
  \put(-5,17){\begin{rotate}{90}\tiny Steepest descent \normalsize\end{rotate}}
  \put(-5,-65){\begin{rotate}{90}\tiny This paper \normalsize\end{rotate}}
 \end{overpic}\;%
\begin{overpic}[width=0.14\linewidth]{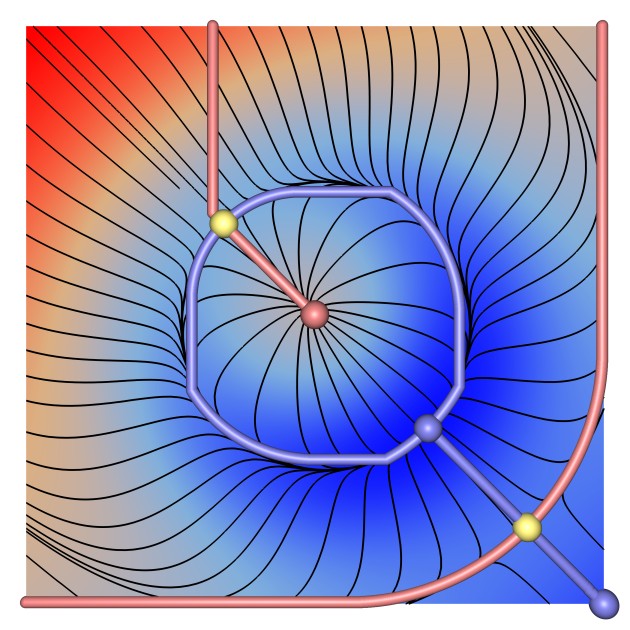} \end{overpic}\;%
\begin{overpic}[width=0.14\linewidth]{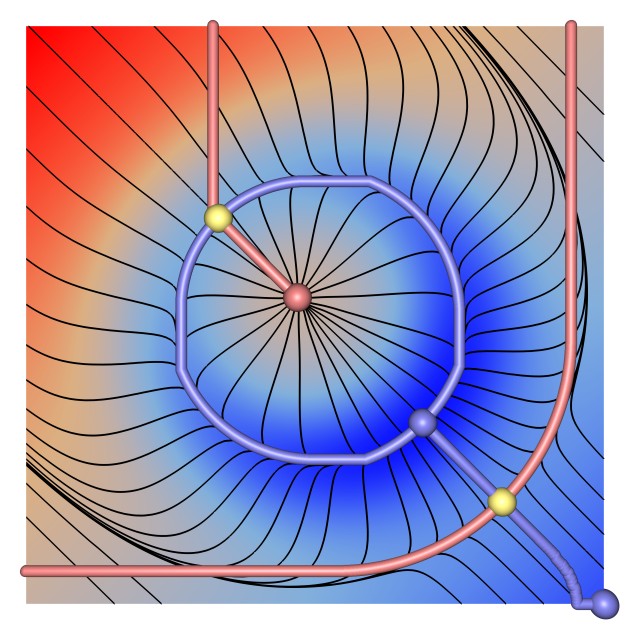} \end{overpic}\;%
\begin{overpic}[width=0.14\linewidth]{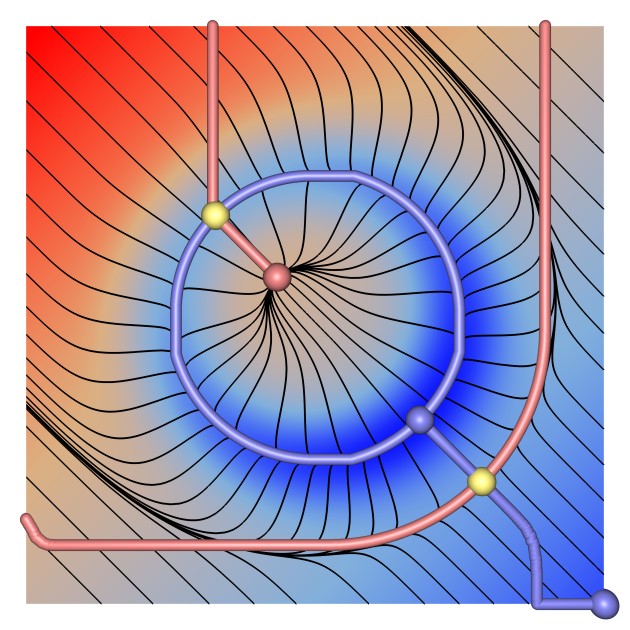} \end{overpic}\;%
\begin{overpic}[width=0.14\linewidth]{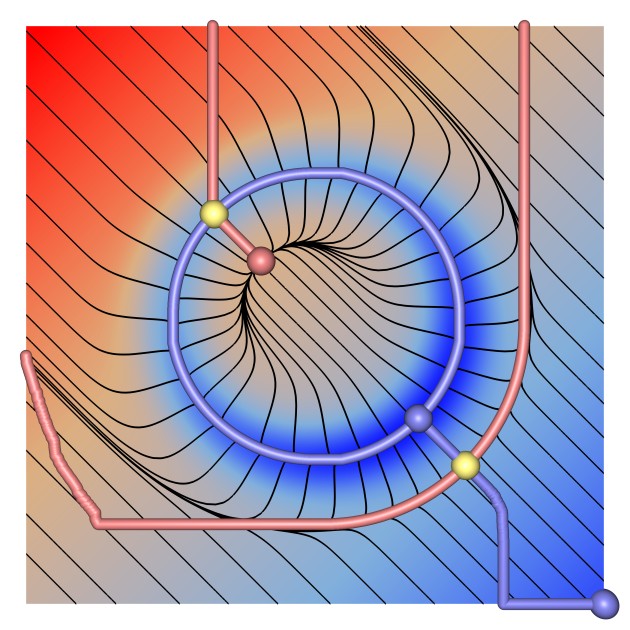} \end{overpic}\;%
\begin{overpic}[width=0.14\linewidth]{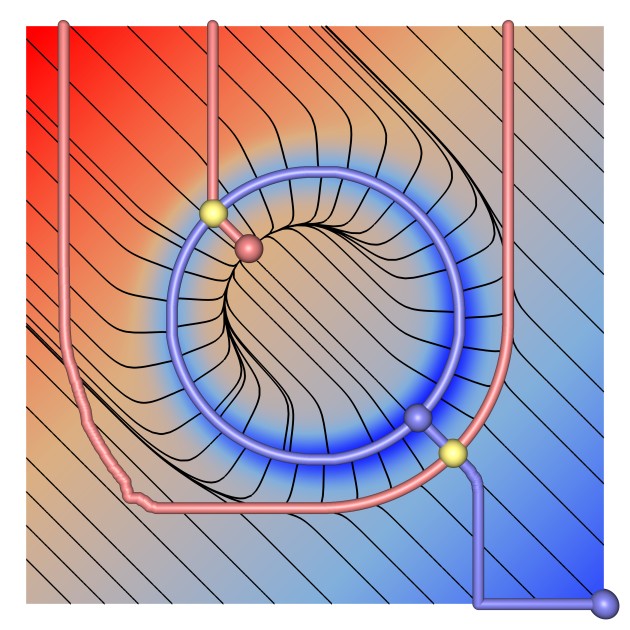} \end{overpic}\vspace{-2ex}\\%
\subfigure[$\alpha=2^0$]{\includegraphics[width=0.14\linewidth]{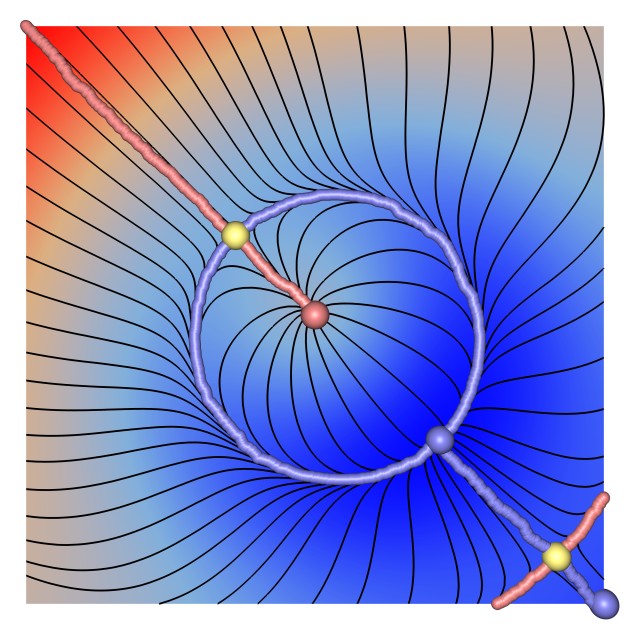}}\;%
\subfigure[$\alpha=2^1$]{\includegraphics[width=0.14\linewidth]{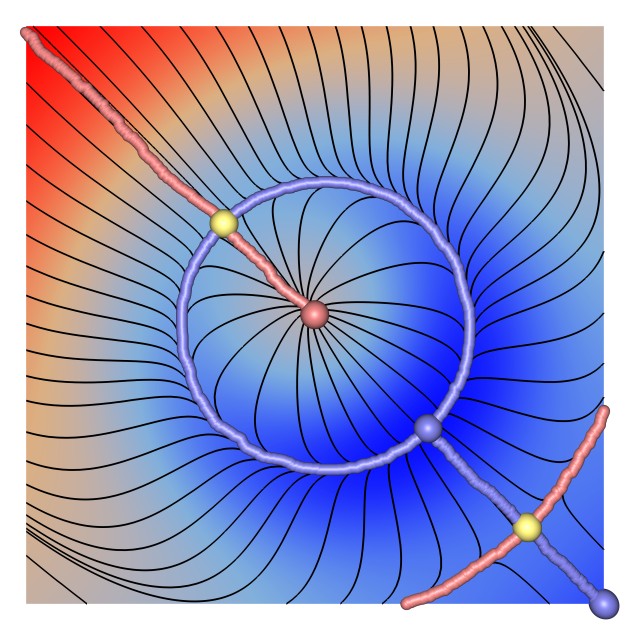}}\;%
\subfigure[$\alpha=2^2$]{\includegraphics[width=0.14\linewidth]{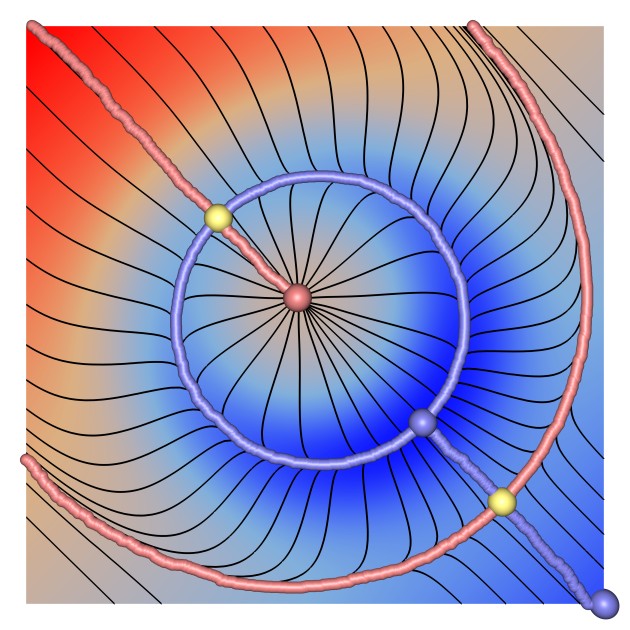}}\;%
\subfigure[$\alpha=2^3$]{\includegraphics[width=0.14\linewidth]{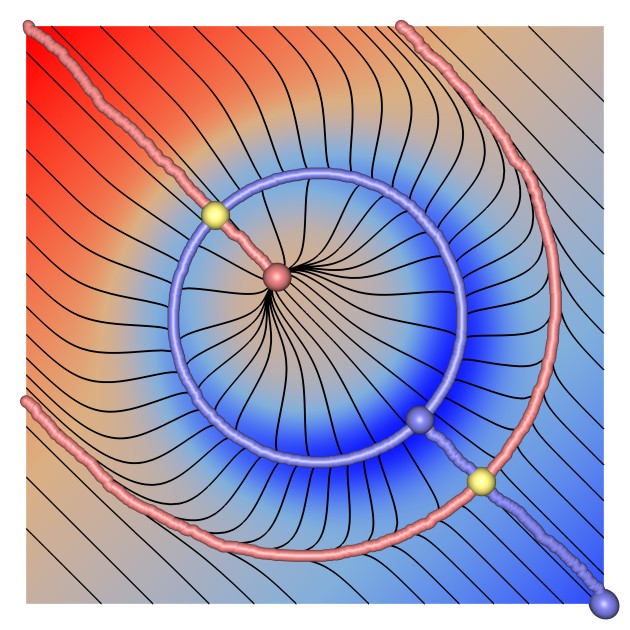}}\;%
\subfigure[$\alpha=2^4$]{\includegraphics[width=0.14\linewidth]{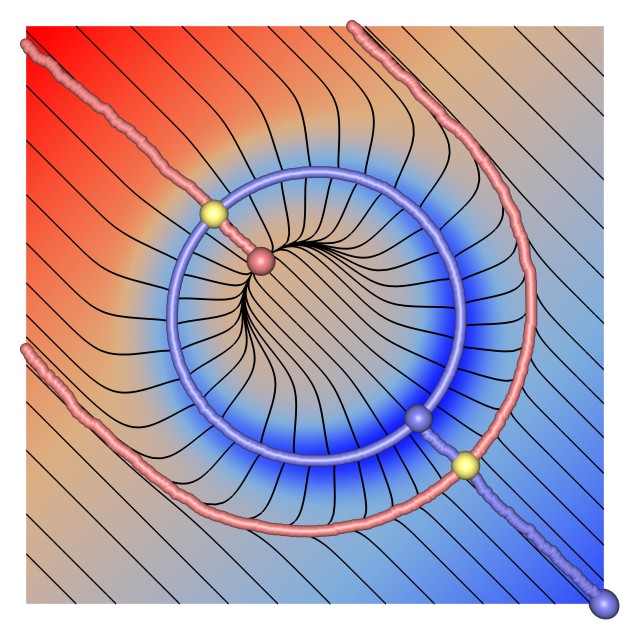}}\;%
\subfigure[$\alpha=2^5$]{\includegraphics[width=0.14\linewidth]{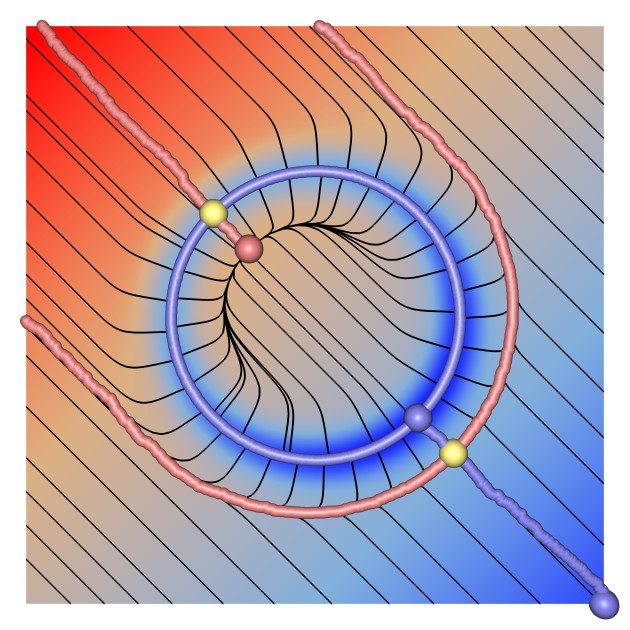}}\\%
\subfigure[Steepest descent error]{\includegraphics[width=0.3725\linewidth]{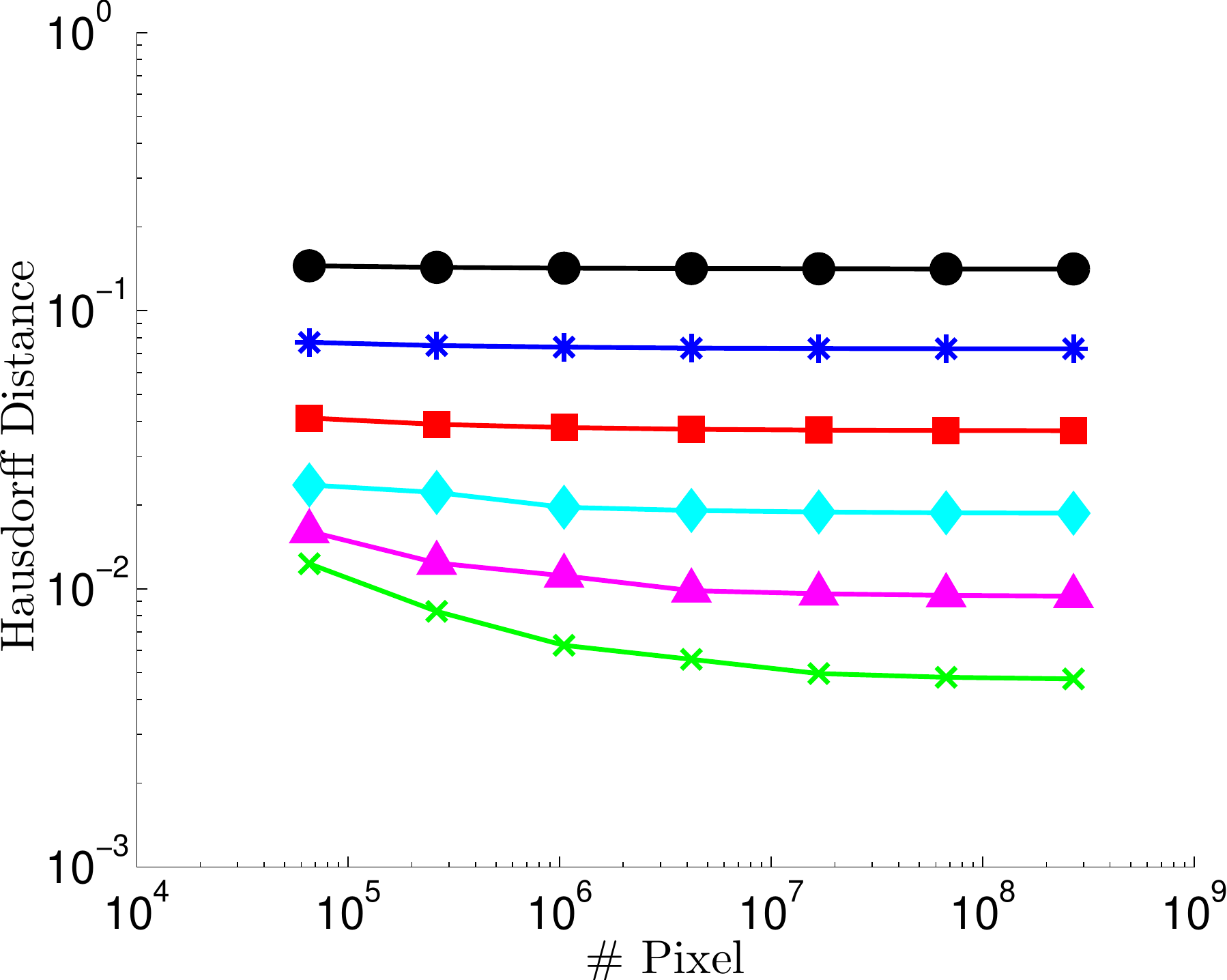}}%\hfill%
\\%
\subfigure[Statistical characteristics of probabilistic error (this paper)]{\includegraphics[width=0.3725\linewidth]{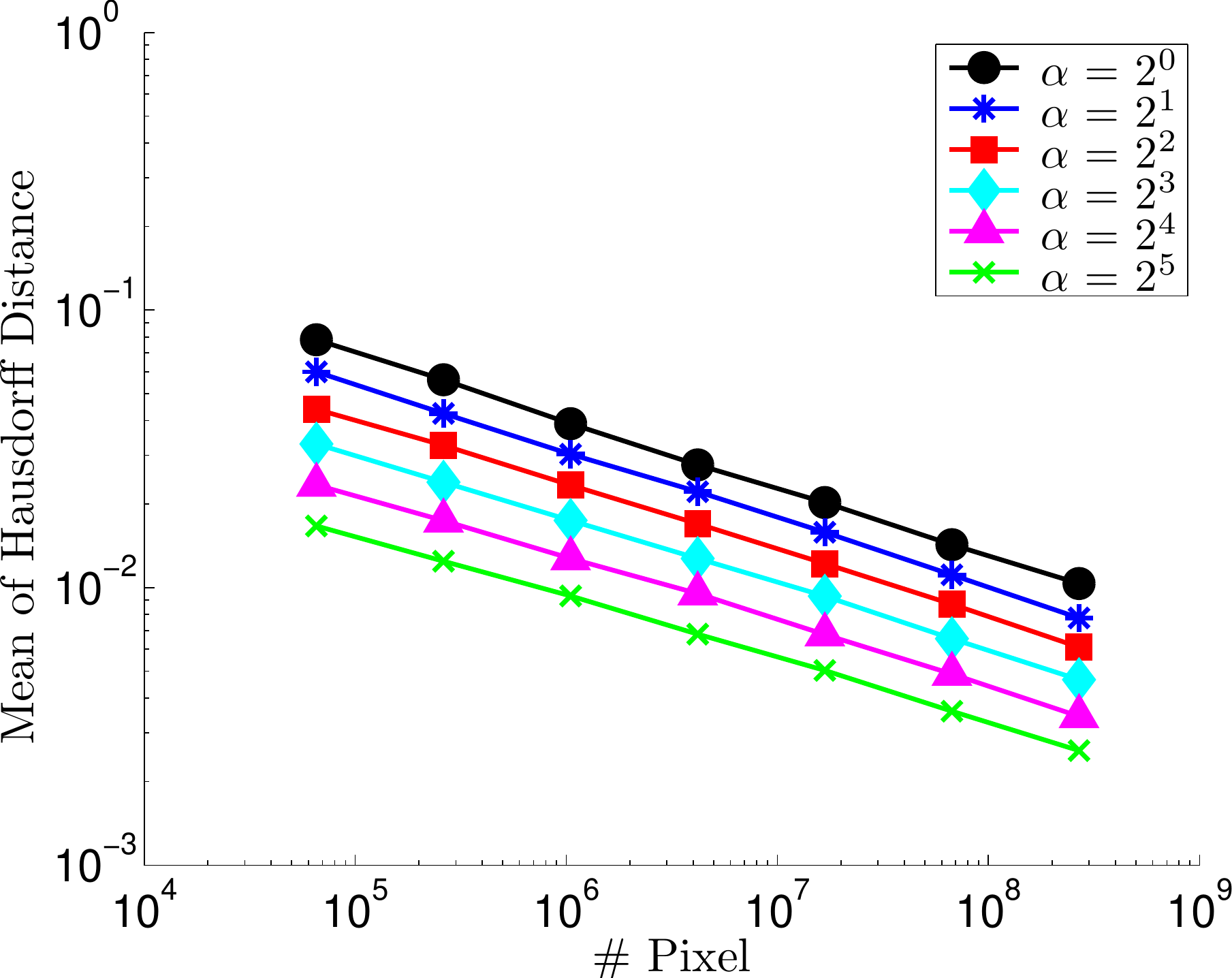}\hspace{8ex}\includegraphics[width=0.3725\linewidth]{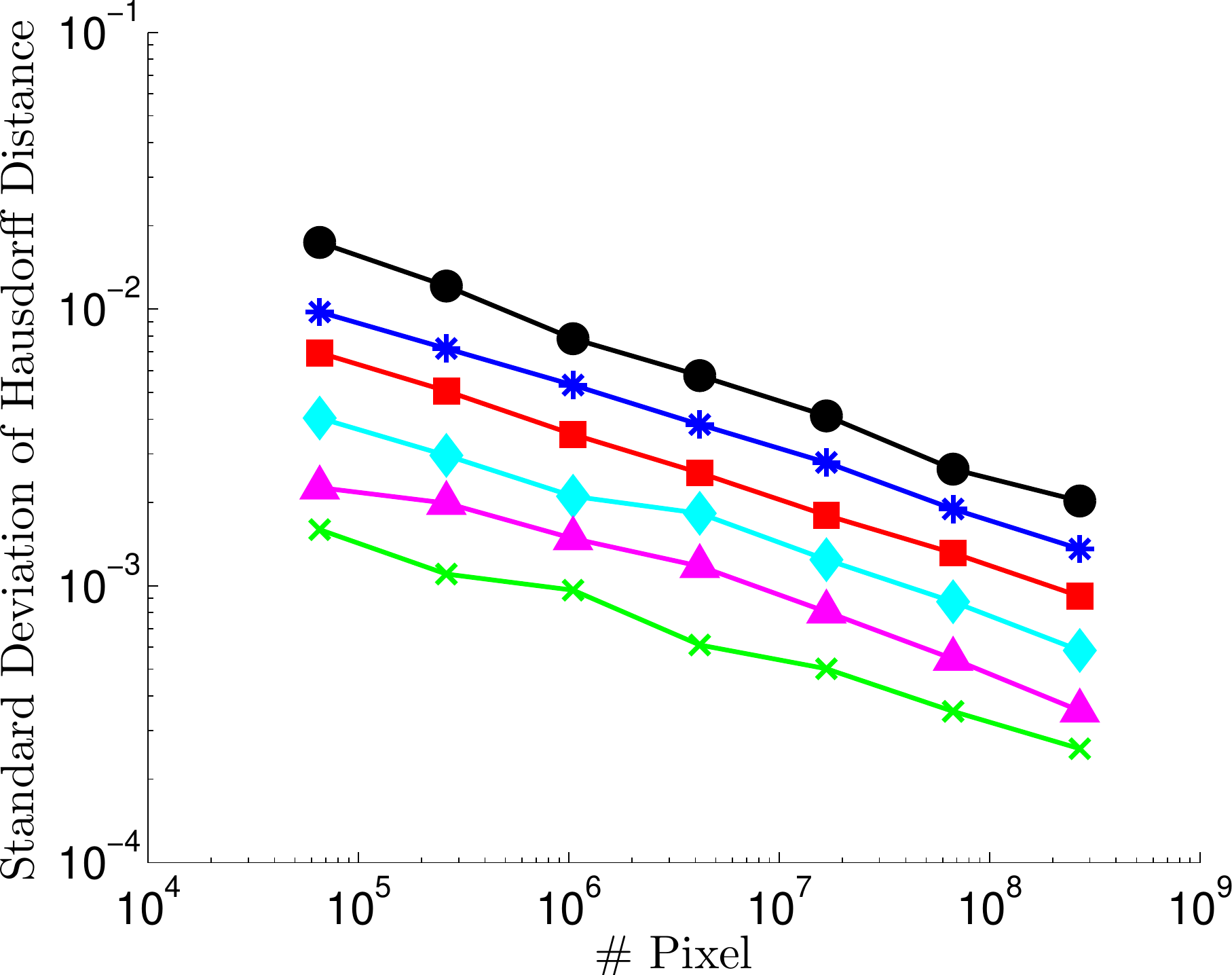}}%
%\subfigure[Standard Deviation (This Paper)]{\includegraphics[width=0.3\linewidth]{figs/ErrorStdDev}}%
\caption{The analytic example $f$ \eqref{eq:analyticFunction}. The first row of (a)-(f) shows the sampled function $f$ color-coded for different choices of $\alpha$. Red denotes a high function value, while blue denotes a low value. Black lines depict integral lines of the gradient $\nabla f$ seeded using the dual streamline technique\c{Rosanwo2009}. The second and third row show the MS-complex based on \alg{alg:CombinatorialGradient} using the steepest descent and our probabilistic edge selection strategy, respectively. Minima, saddles and maxima are shown as blue, yellow and red spheres, whereas the 0- and 1-separatrices are shown as blue and red lines, respectively. Since separatrices are integral lines of the gradient, the blue and red lines should follow the black lines. (g) and (h) show the Hausdorff distance of the approximated center circle (blue) to the reference circle (white, first row) for different choices of $\alpha$ and increasing resolution. (g) shows the evolution of the error using the steepest descent approach. (h) shows two statistical characteristics of the error of the probabilistic approach proposed in this paper.}%
\label{fig:error}%
\end{figure*}

\paragraph{An analytic function.} 
Let $\Omega=[-2,2]^2$ and $\alpha\in\mathbb{R}^+$. The function $f:\,\Omega\rightarrow\mathbb{R}$ is given as
\begin{equation}
f(x,y) = -e^{-\alpha\,\left(\sqrt{x^2+y^2}-1\right)^2}-0.3\,(x+y).
\label{eq:analyticFunction}
\end{equation}
The function $f$ describes a circle engraved on a tilted plane. The sharpness of this circle is defined by the parameter $\alpha$. For $\alpha\rightarrow\infty$, the circle becomes arbitrary sharp. For $\alpha\rightarrow0$, $f$ gets flattened. Varying $\alpha$ allows us to simulate smooth as well as sharp features appearing in many applications. An illustration of $f$ sampled on a $2048^2$ grid for different choices of $\alpha$ is given in the first row of \fig{fig:error}. Integral lines of the continuous gradient $\nabla f$ are depicted by black lines using the dual streamline seeding technique\c{Rosanwo2009}. Converging integral lines indicate thereby the existence of a separatrix. In the following paragraphs, we choose the engraved circle as a reference feature. For illustration, it is shown as a white circle line in the first row of \fig{fig:error}.

\paragraph{A qualitative comparison.}
We applied \alg{alg:CombinatorialGradient} using the steepest descent version as well as our probabilistic version to construct a combinatorial gradient of $f$ for different choices of $\alpha$. The resulting MS-complexes of the steepest descent version are shown in the second row of \fig{fig:error}. Our reference feature -- the white circle -- is visually well recovered for large values of $\alpha$, which confirms also the extraction results of recently proposed methods\c{Cazals:SCG-03,Kasten_VRG,Weinkauf2009}. In many applications, the desired features are sufficiently sharp.

However, deviations to the reference circle become visible if the feature gets smooth, i.e., for small choices of $\alpha$. The steepest descent version of \alg{alg:CombinatorialGradient} is not able to recover the circle for $\alpha=2^0$ and $\alpha = 2^1$. The probabilistic approach, in contrast, is able to recover the circle for all choices of $\alpha$. The resulting MS-complexes are shown in the third row of \fig{fig:error}.

\begin{figure}[htb]%
\centering%
\includegraphics[width=0.55\linewidth]{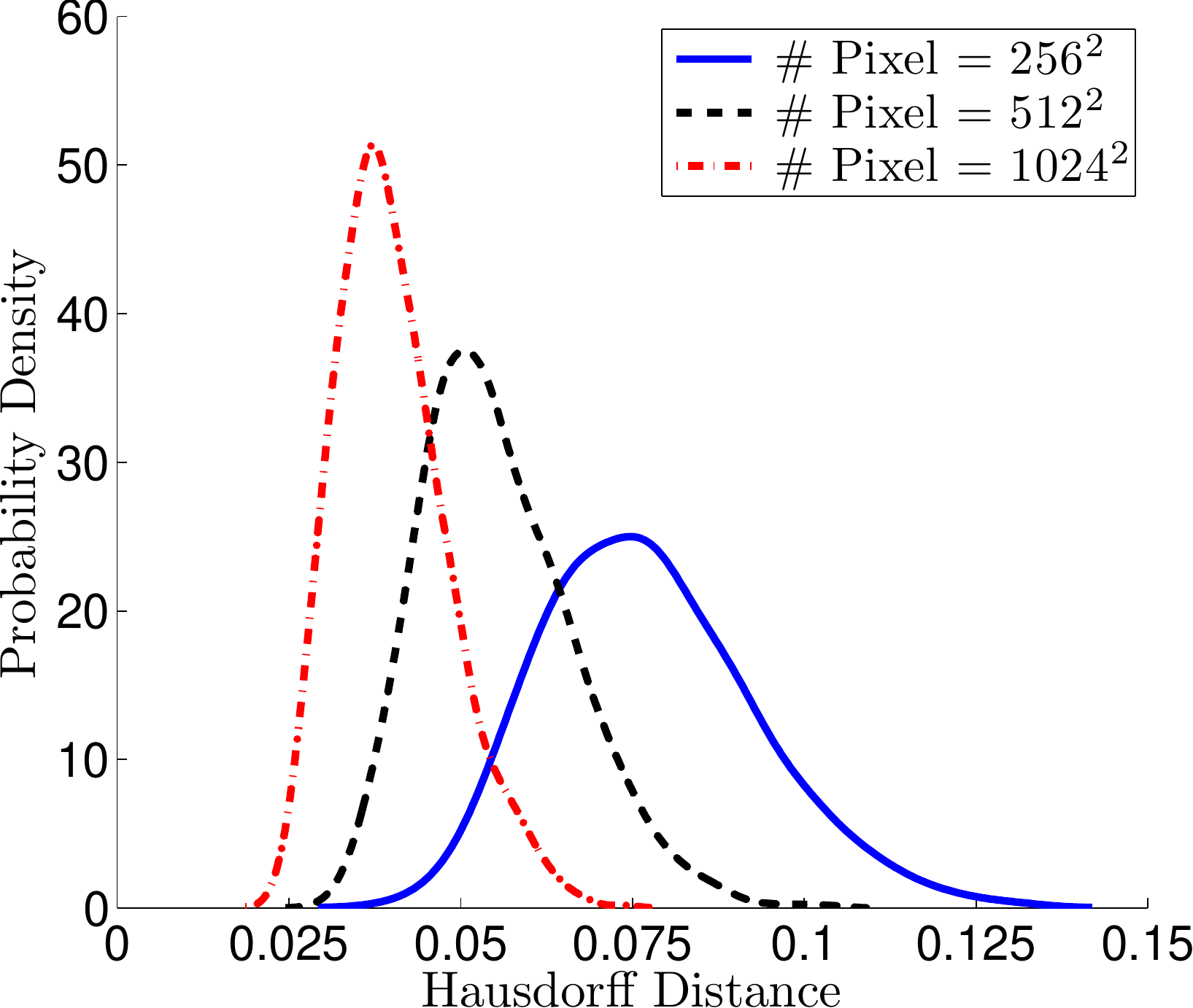}
%\subfigure[Sharpness 0]{\includegraphics[width=0.3\linewidth]{figs/HausdorffDistance256}}\hfill%
%\subfigure[Sharpness 0]{\includegraphics[width=0.3\linewidth]{figs/HausdorffDistance512}}\hfill%
%\subfigure[Sharpness 0]{\includegraphics[width=0.3\linewidth]{figs/HausdorffDistance1024}}\\%
%\subfigure[Sharpness 0]{\includegraphics[width=0.3\linewidth]{figs/HausdorffDistance256}}\hfill%
%\subfigure[Sharpness 0]{\includegraphics[width=0.3\linewidth]{figs/HausdorffDistance512}}\hfill%
%\subfigure[Sharpness 0]{\includegraphics[width=0.3\linewidth]{figs/HausdorffDistance1024}}

\caption{Distribution of Hausdorff distance error. The blue, black and red curves show the estimated probability density function  of the Hausdorff distance between the center circle and the reference circle (as shown in \fig{fig:error}) for different resolutions.}%
\label{fig:kernelDensity}%
\end{figure}

\begin{figure}[!htp]%
\centering%
\subfigure[Input image]{\includegraphics[width=0.4\linewidth]{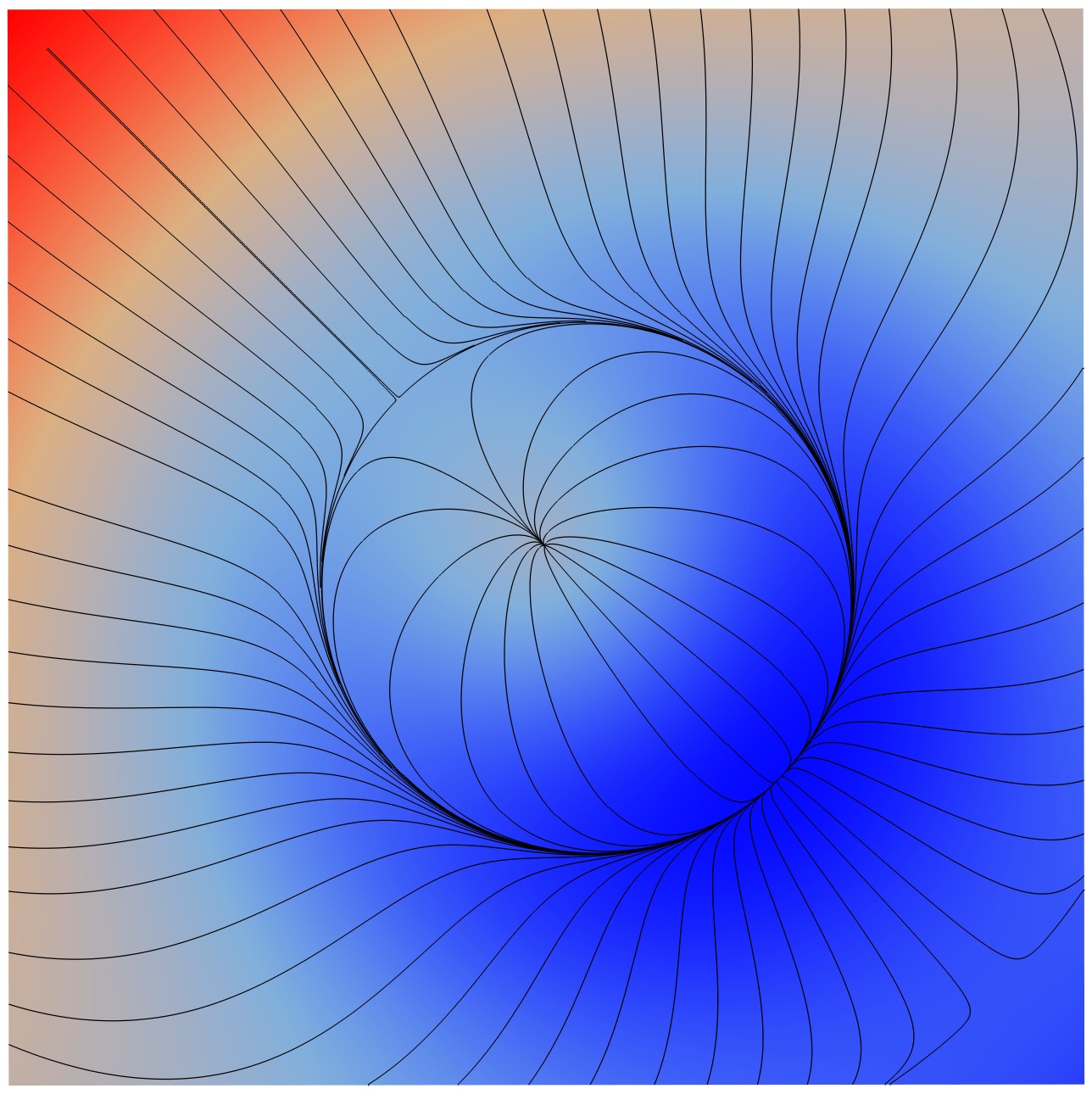}}\hspace{0.5cm}%
\subfigure[Steepest descent]{\includegraphics[width=0.4\linewidth]{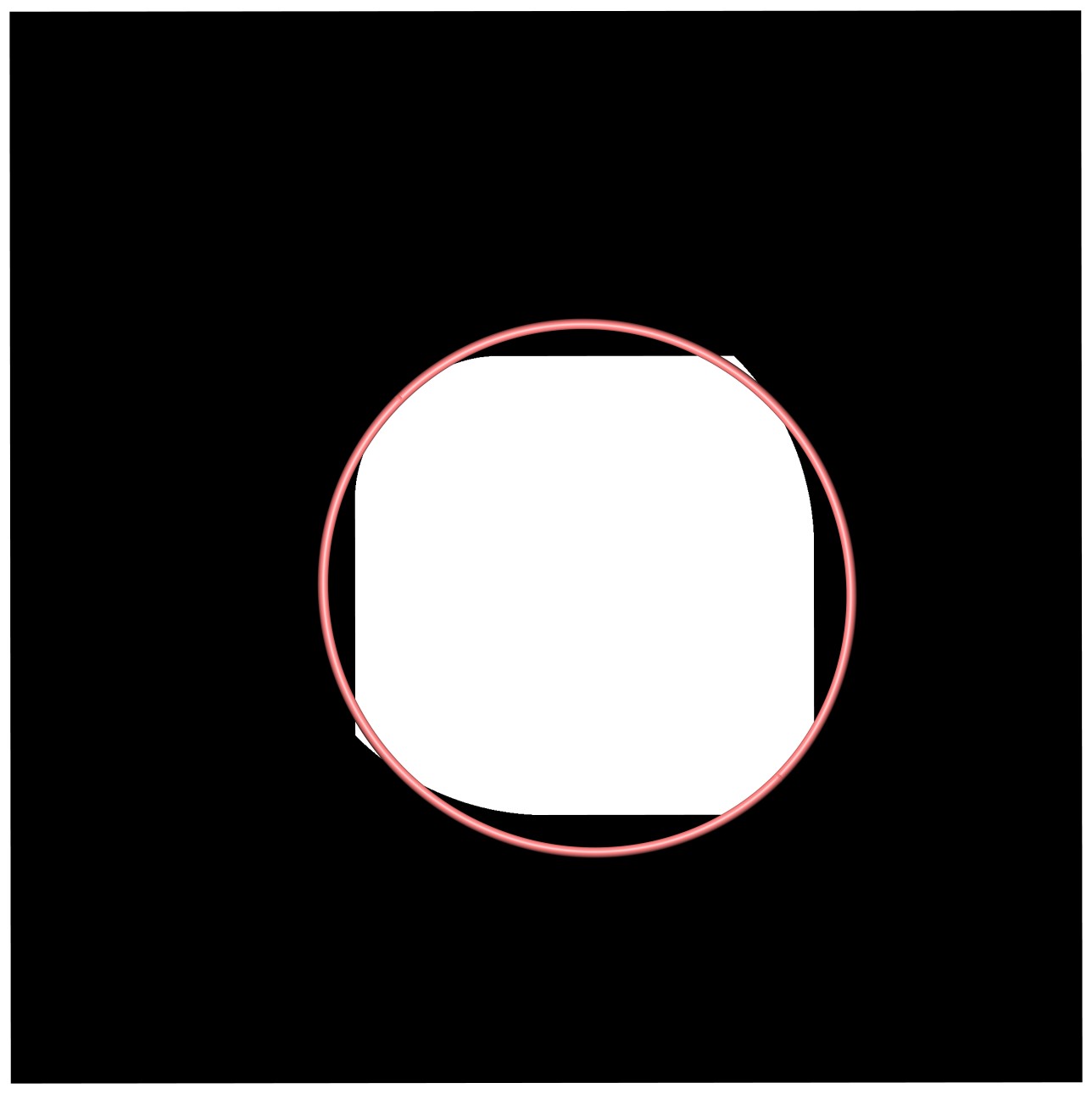}}\\%
\subfigure[Watershed 4-connectivity]{\includegraphics[width=0.4\linewidth]{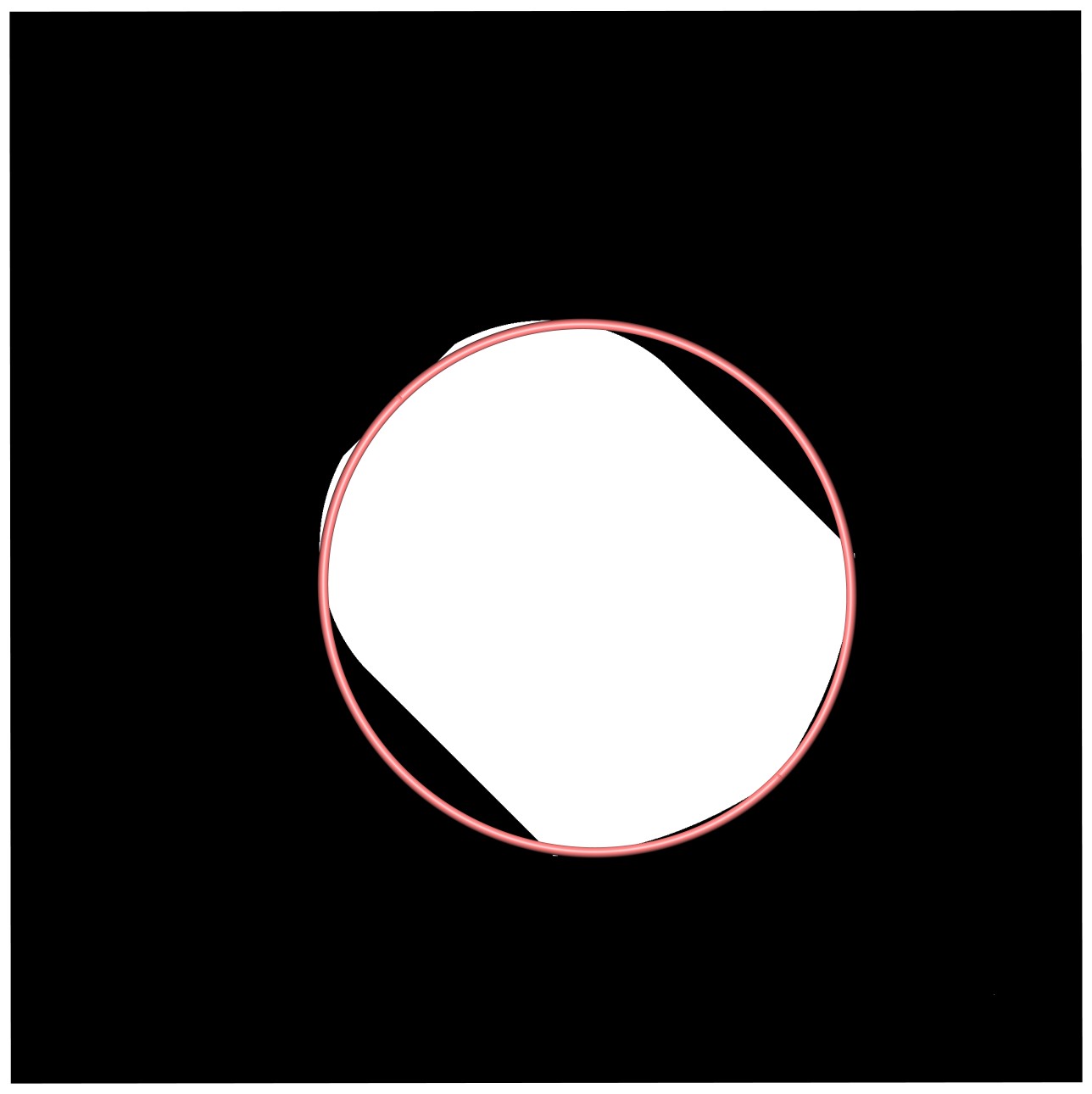}}\hspace{0.5cm}%
\subfigure[Watershed 8-connectivity]{\includegraphics[width=0.4\linewidth]{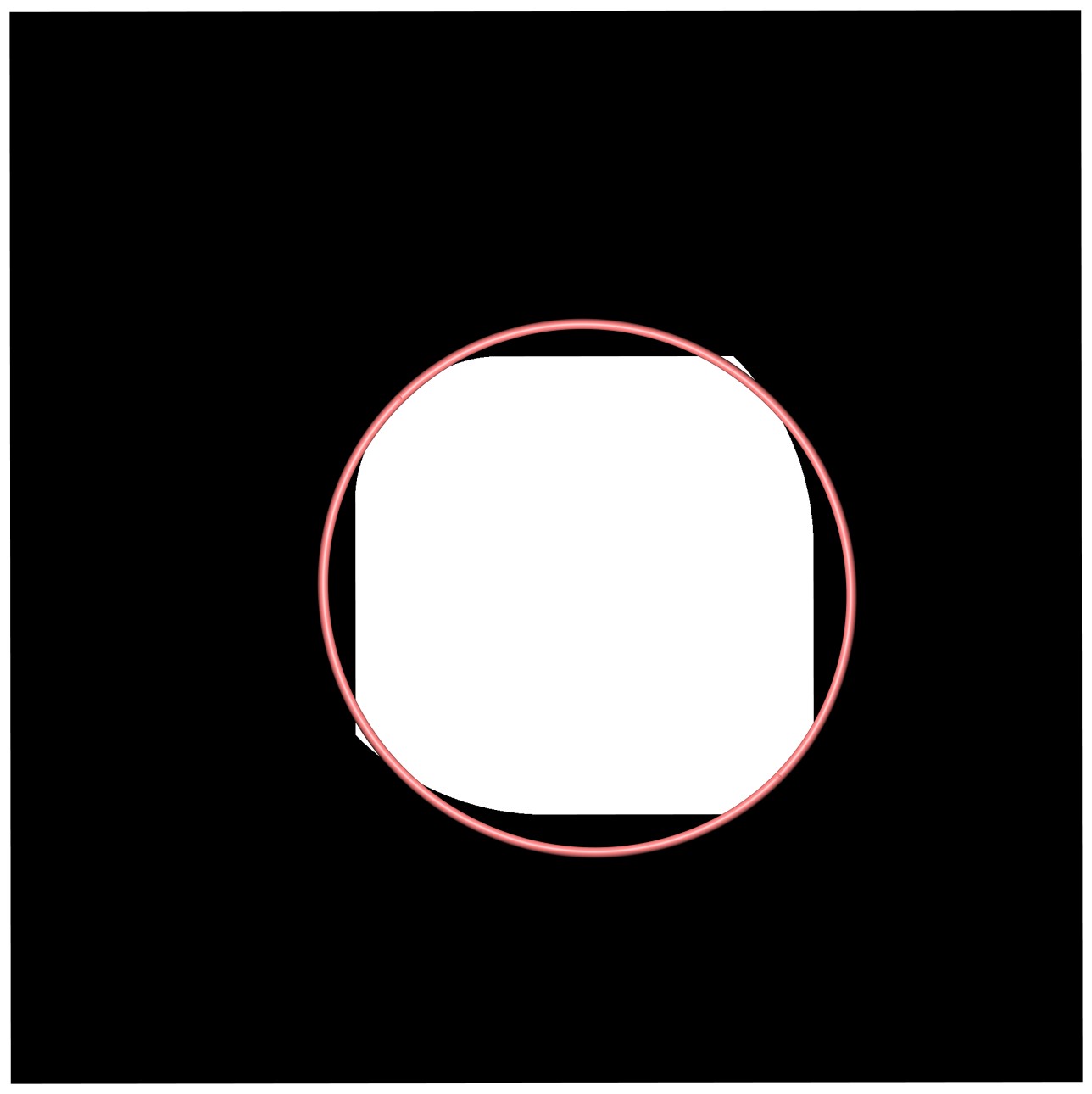}}\\%
\subfigure[Spanning forest (triangulated)]{\includegraphics[width=0.4\linewidth]{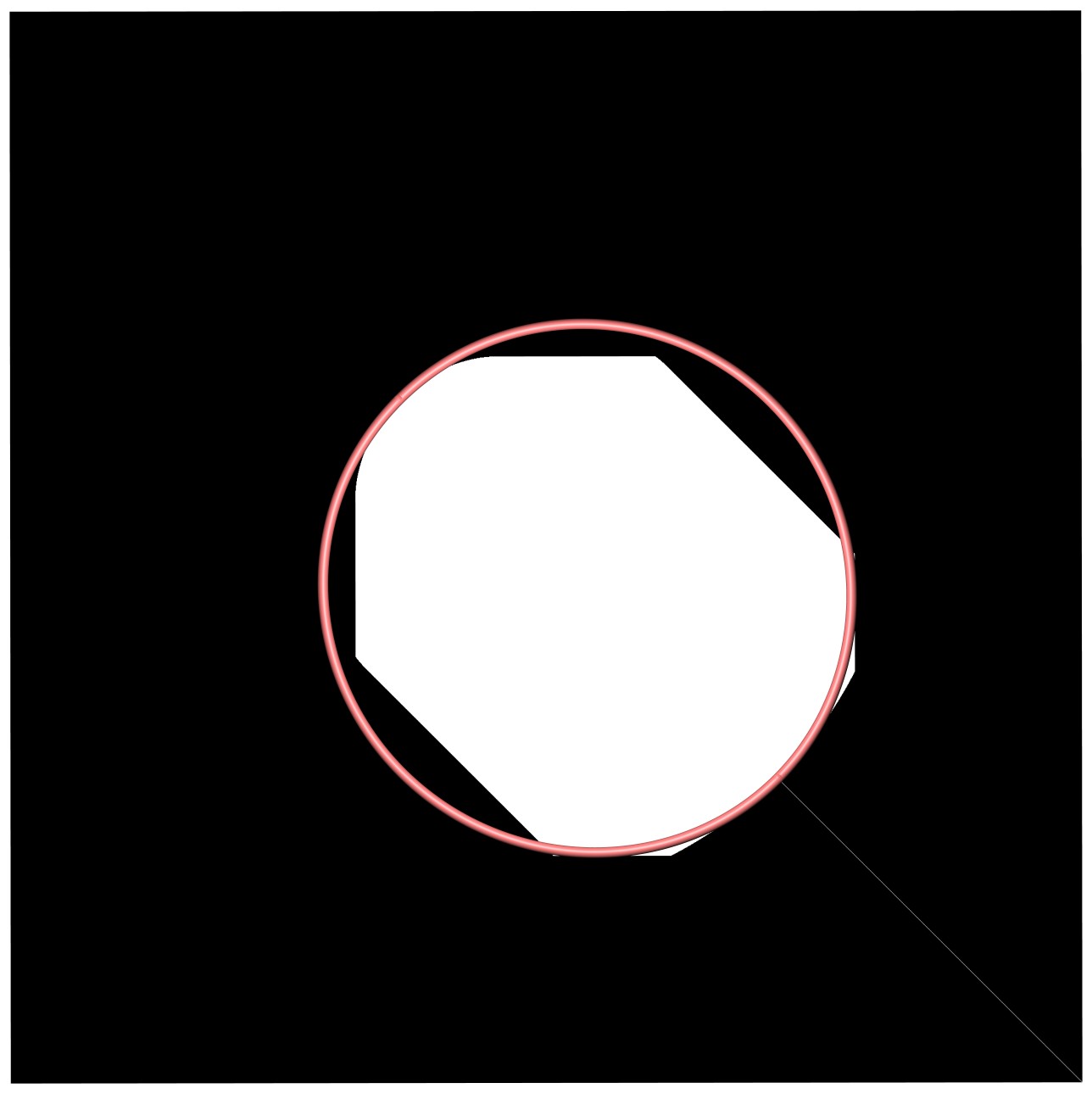}}\hspace{0.5cm}%
\subfigure[This paper]{\includegraphics[width=0.4\linewidth]{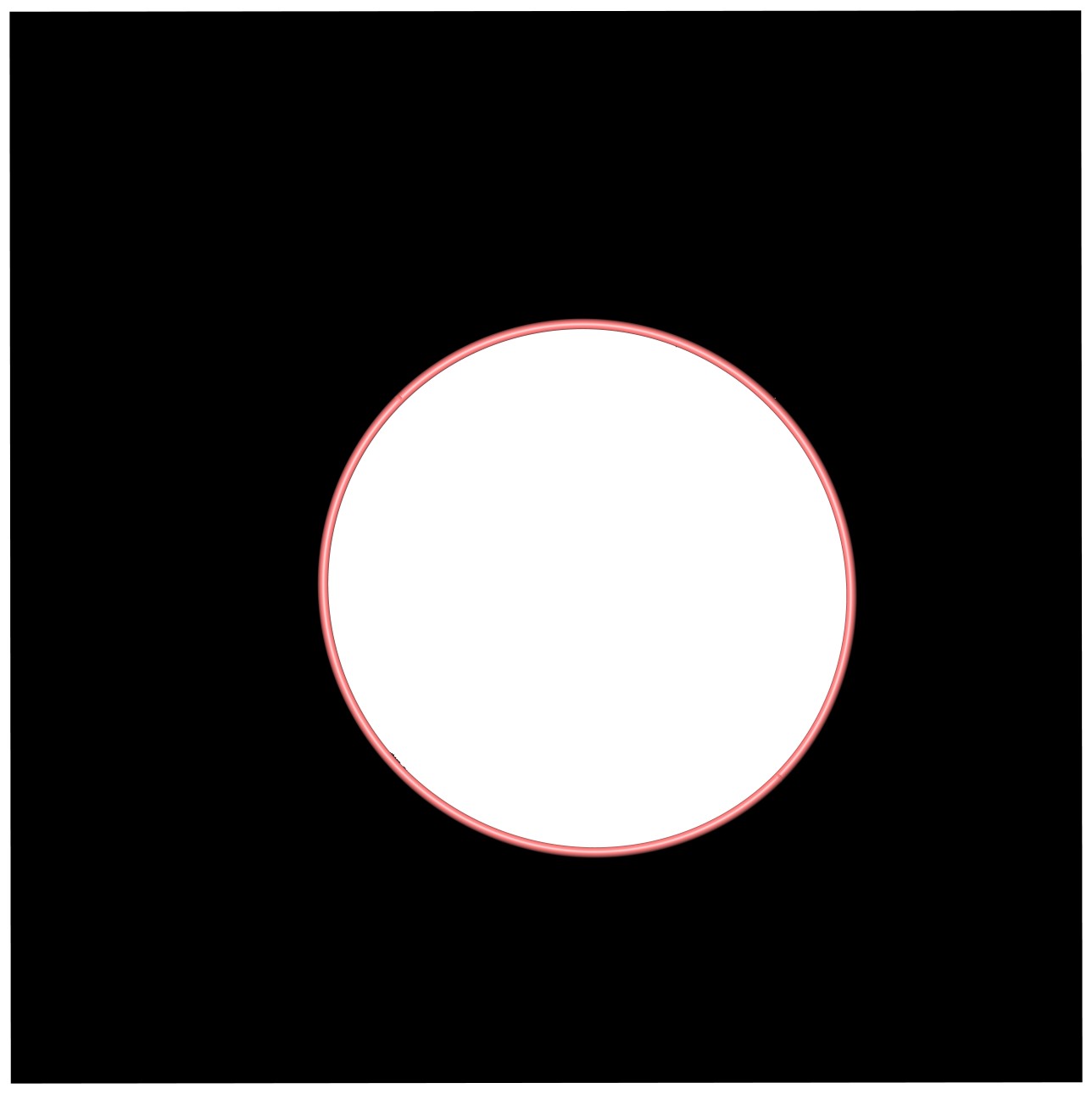}}%
\caption{Segmentation comparison. The analytic function $f$ as defined in \eqref{eq:analyticFunction} with $\alpha=1$ is visualized as in \fig{fig:error} -- the black lines depict integral lines of the gradient. Converging integral lines indicate the separatrices/watersheds. (b) shows the result using \alg{alg:CombinatorialGradient} with the steepest descent strategy. (c) and (d) show the result using the Matlab implementation of the classic watershed algorithm\c{Meyer1994}. (e) shows the result using a fitted spanning forest approach\c{Cazals:SCG-03}. (f) shows the segmentation using our probabilistic approach based on \alg{alg:CombinatorialGradient}. The red circle shows the reference separatrix/watershed of the continuous function.}%
\label{fig:watershed}%
\end{figure}

\paragraph{A quantitative comparison.}
To quantify the approximation error, we measured the Hausdorff distance\c{Hausdorff1914} of the reference circle to the approximated circle. \fig{fig:error} g) shows the approximation error for \alg{alg:CombinatorialGradient} using the steepest descent strategy in a log-log plot. Although the extraction result looked visually reasonable for large $\alpha$-values, there is no convergence for any $\alpha$. The Hausdorff distance to the reference circle does not decrease when a finer grid is employed.

For the probabilistic approach, we did $200$ runs of \alg{alg:CombinatorialGradient} using the method presented in \sec{sec:method_method}. \fig{fig:error} h) shows the mean value of the Hausdorff distance and its standard deviation. Both quantities are converging to zero. Hence, the sampling as well as the quantization error mentioned in \sec{sec:method_method} are reduced when a finer grid is employed. However, it needs to be noted that the approximation error for very sharp features ($\alpha=2^5$) is slightly larger compared to the steepest descent strategy when coarse grids are used.

In \fig{fig:kernelDensity}, we plotted a kernel density estimate\c{Botev2010} of the probability density of the Hausdorff distance error for different resolutions. We applied \alg{alg:CombinatorialGradient} using the probabilistic approach $1000$ times to obtain a sufficient number of samples for the density estimate. The overall shape of the densities suggests a sequence of binomial error distributions that converge to a normal distribution.

%While the width of the densities is getting smaller as the resolution is increased, their mean value tends to zero. This nicely confirms the observations of \fig{fig:error}. The overall shape of the distributions suggests a binomial error distribution.

\paragraph{A segmentation comparison.}
In \fig{fig:watershed} we compared ourselves (f) with different established segmentation approaches: steepest descent\c{Robins2011} (b), Meyer's watershed algorithm\c{Meyer1994} using a 4- (c) and 8-connectivity (d) provided by the image processing toolbox of Matlab\c{MATLAB2009}, and a fitted spanning forest approach\c{Cazals:SCG-03} (e). Since the implementation of\c{Cazals:SCG-03} that was available to us requires a triangulation we subdivided each pixel into two triangles. 

The objective is the segmentation of the inner part of our reference circle of $f$ with $\alpha=1$ sampled on a $2048^2$ grid. It can be observed that the steepest descent strategy yields a similar result as the watershed algorithm using an 8-connectivity, while the result of the spanning forest approach looks similar to the watershed result using a 4-connectivity. However, none of these approaches is able to segment the circle, in contrast, to the probabilistic approach presented in this paper.

\begin{figure}[!b]%
\centering%
\subfigure[Steepest descent ($256\times 4096$)]{\includegraphics[width=0.4\linewidth]{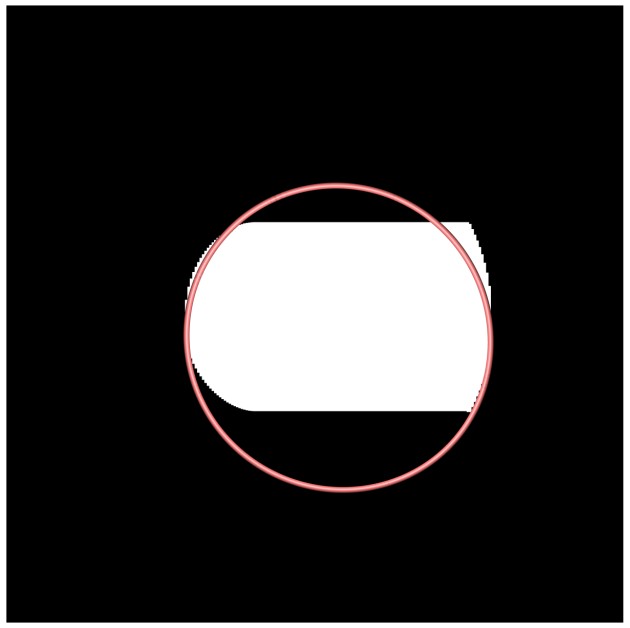}}\hspace{0.5cm}%
\subfigure[This paper ($256\times 4096$)]{\includegraphics[width=0.4\linewidth]{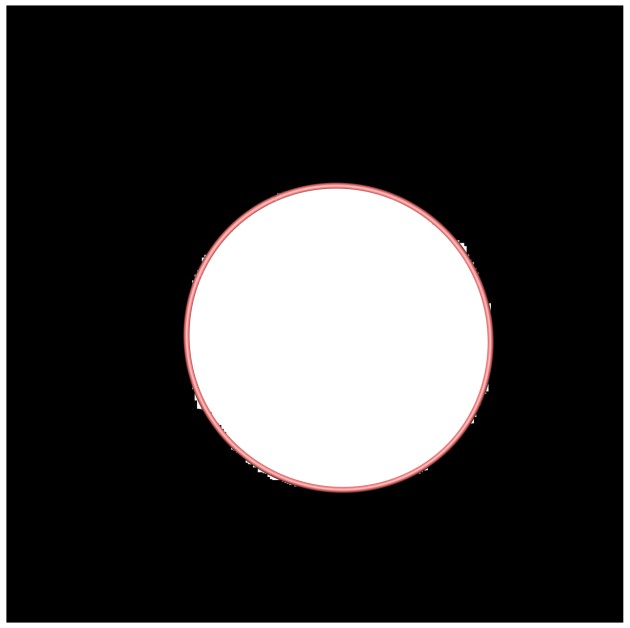}}\\%
\subfigure[Steepest descent ($2048\times 4096$)]{\includegraphics[width=0.4\linewidth]{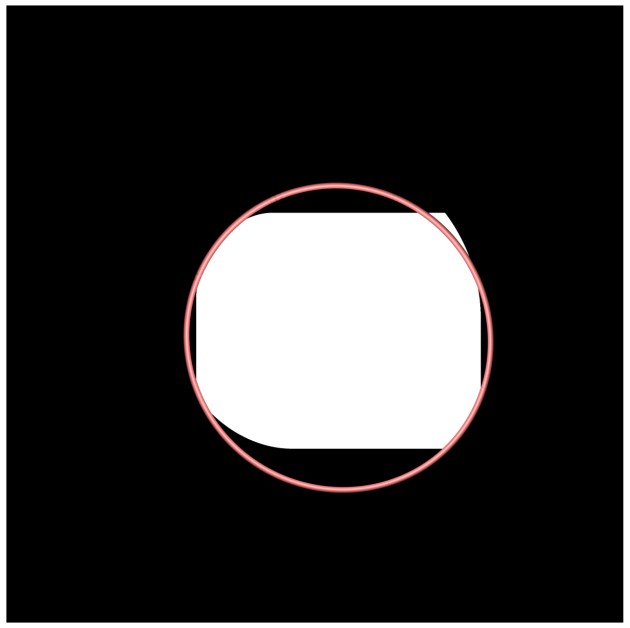}}\hspace{0.5cm}%
\subfigure[This paper ($2048\times 4096$)]{\includegraphics[width=0.4\linewidth]{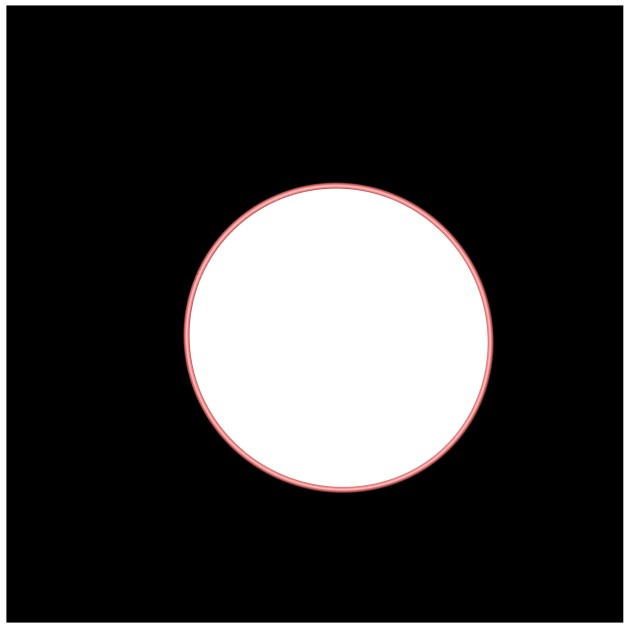}}%
\caption{Segmentation comparison on anisotropic grids. The left column shows the segmentation of the analytic function $f$ \eqref{eq:analyticFunction} with $\alpha=1$ using \alg{alg:CombinatorialGradient} with the steepest descent approach on two anisotropic grids. The right column shows the corresponding results for our probabilistic approach. The red circle depicts the continuous separatrix/watershed of $f$.}%
\label{fig:anisotropy}%
\end{figure}
\paragraph{Anisotropic grids.}
In certain applications, the image data is not provided on uniform grids. Based on the imaging technique anisotropic grids might be employed. In \fig{fig:anisotropy}, we illustrate the behavior of the steepest descent and our probabilistic strategy applied to $f$ with $\alpha=1$ using two different anisotropic grids. 

It can be seen in \fig{fig:anisotropy} a) and c) that the result of the steepest descent approach heavily depends on the kind of anisotropy. The resulting segmentation is distorted when the anisotropy becomes large. However, this is not the case for our probabilistic approach, see \fig{fig:anisotropy} b) and d). 

\begin{figure}%
\centering%
\subfigure[Ground truth]{\begin{overpic}[width=0.3\linewidth]{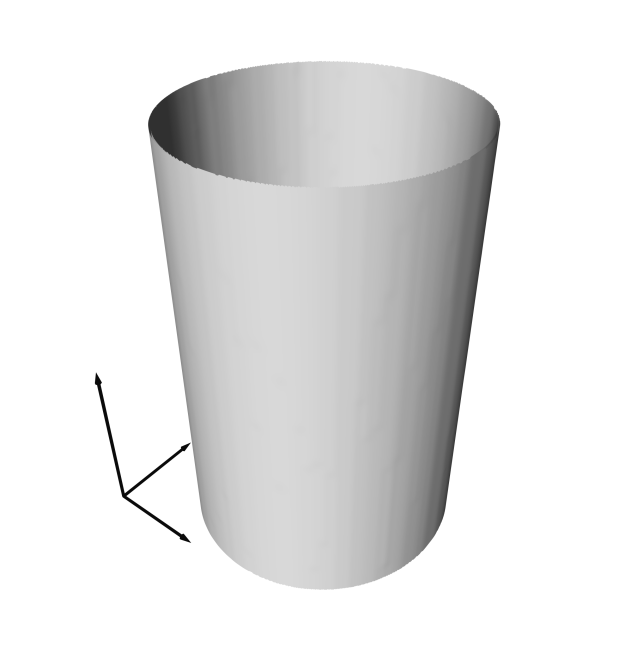}
\put(20,10){\footnotesize$x$}
\put(22,31){\footnotesize$y$}
\put(10,31){\begin{rotate}{27}\footnotesize$\theta$\end{rotate}}
\end{overpic}}\hfill%
\subfigure[Steepest descent]{\includegraphics[width=0.3\linewidth]{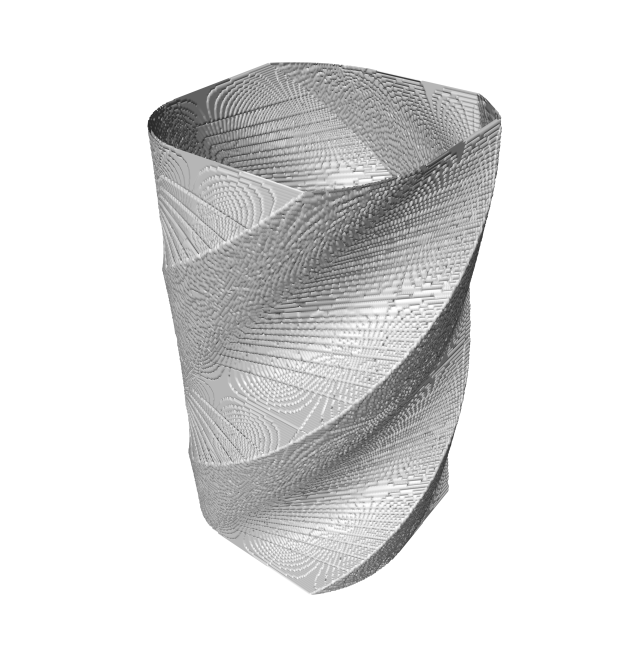}}\hfill%
%\subfigure[Greedy Smooth]{\includegraphics[width=0.19\linewidth]{figs/greedy_smooth}}\hfill%
\subfigure[This paper]{\includegraphics[width=0.3\linewidth]{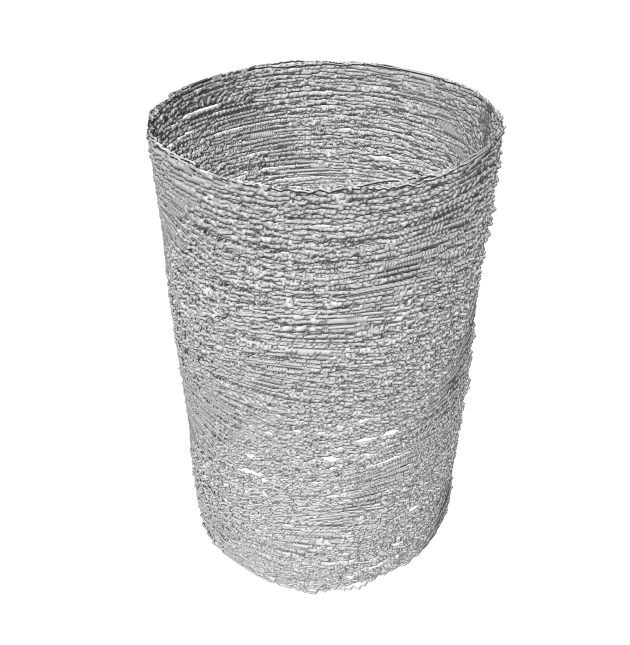}}%
%\subfigure[Our Method Smooth]{\includegraphics[width=0.19\linewidth]{figs/probalistic_smooth}}%
\caption{Grid dependence. (a) shows the ground truth of the rotating function $f_\theta$ as defined in \eqref{eq:rotatedFunction}. The gray surface depicts the evolution of the spatial reference feature shown in \fig{fig:error}. (b) and (c) show the discrete result of \alg{alg:CombinatorialGradient} using the steepest descent edge selection and our probabilistic strategy, respectively.}%
\label{fig:timeDependent}%
\end{figure}

%\newpage
%\textbf{Time varying data.}
\paragraph{Grid dependency.}
To evaluate the grid dependency of our approach, we rotated $f$ by different angles $\theta\in[\nicefrac{-\pi}{2},\nicefrac{\pi}{2}]$. Since this is a rigid transformation of $f$, a rotation by $-\theta$ of the extracted separatrices should coincide with the the original separatrices of $f$.
%we sampled a rotated version of $f$ on a uniform grid. 
%The extraction result is then rotated back to the original frame of reference. 
%There are time dependent data sets where the spatial separatrices indicate relevant features\c{Kasten_VRG}. We now investigate the behavior of our method for this kind of data. We applied it to a rotating version of $f$ as follows. 
The rotated function $f_\theta:\,\Omega\times[\nicefrac{-\pi}{2},\nicefrac{\pi}{2}]\rightarrow\mathbb{R}$ is given by 
\begin{equation}
f_\theta(x,y) = f(R(x,y,\theta))
\label{eq:rotatedFunction}
\end{equation}
with
\begin{equation*}
R(x,y,\theta)=\left(
\begin{array}{cc}
  \cos(\theta) & -\sin(\theta)\\ 
  \sin(\theta) & \cos(\theta)
\end{array}
\right)
\left(
\begin{array}{c}
x\\
y
\end{array}
\right).
\end{equation*}
We concentrate ourselves on the reference feature of Figure~\ref{fig:error}. Since the center circle is rotated and back-rotated by $\theta$, the ground truth is a cylinder. We extracted the center circle using the steepest descent and our probabilistic strategy for each rotation step. The spatial domain was discretized using a $1024^2$ grid and a full rotation was sampled using $1024$ steps.

In order to obtain a segmentation of this region over the the range of the rotation, we stacked the spatial separatrices on top of each other and connected them to a surface. The result is shown in \fig{fig:timeDependent}. %The ground truth is generated using a numerical approach using autonomous ordinary differential equations\c{Weinkauf2008a}. 

The result of the steepest descent strategy shown in \fig{fig:timeDependent} b) clearly reflects the structure of the grid. The surface is snapped to the grid, no cylindrical shape can be observed. In contrast, the probabilistic approach shown in \fig{fig:timeDependent} c) is a rough approximation of the ground truth, and the cylindrical shape is well recovered. 

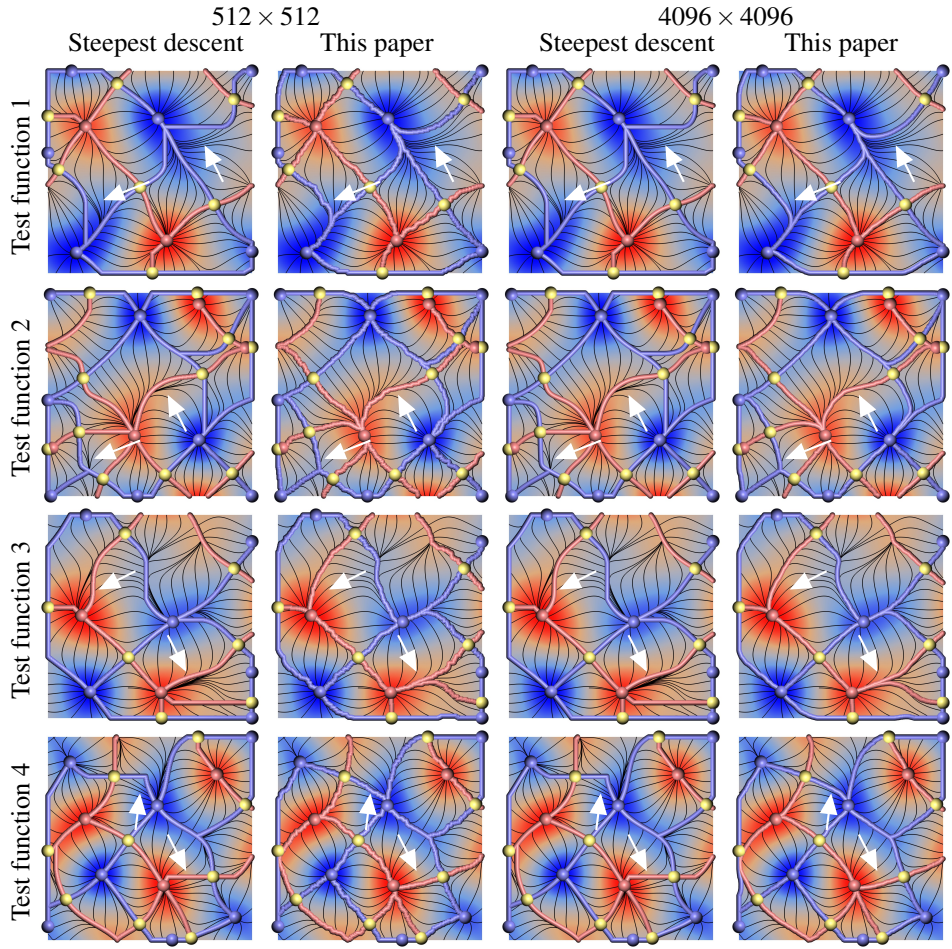
\begin{figure*}[!t]%
\centering%
\vspace{4ex}
\pgfsetxvec{\pgfpoint{\linewidth}{0}}%
\pgfsetyvec{\pgfpoint{0}{\linewidth}}%
\begin{tikzpicture}[>=latex',join=bevel,]%
\node (text1) at (-1,0.03) {\begin{rotate}{90}Test function $1$\end{rotate}};
\node[inner sep=0pt,above right] (pic1) at (-0.99,0) {\begin{overpic}[width=0.24\linewidth]{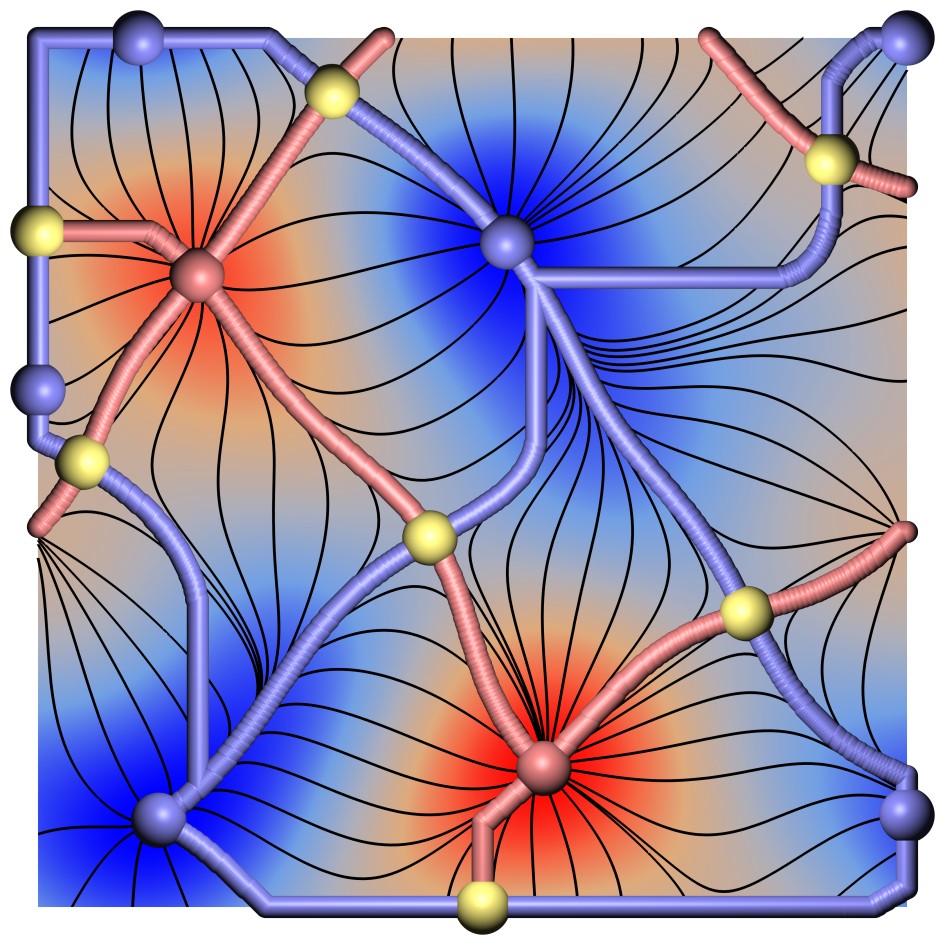} 
  \fontsize{10}{10}
  \put(78,117){$512\times 512$}
  \put(13,105){Steepest descent}
  \put(127,105){This paper}
  \put(280,117){$4096\times 4096$}
  \put(226,105){Steepest descent}
  \put(337,105){This paper}
 \end{overpic}};\;%
\node[inner sep=0pt,above right] (pic2) at (-0.74,0) {\includegraphics[width=0.24\linewidth]{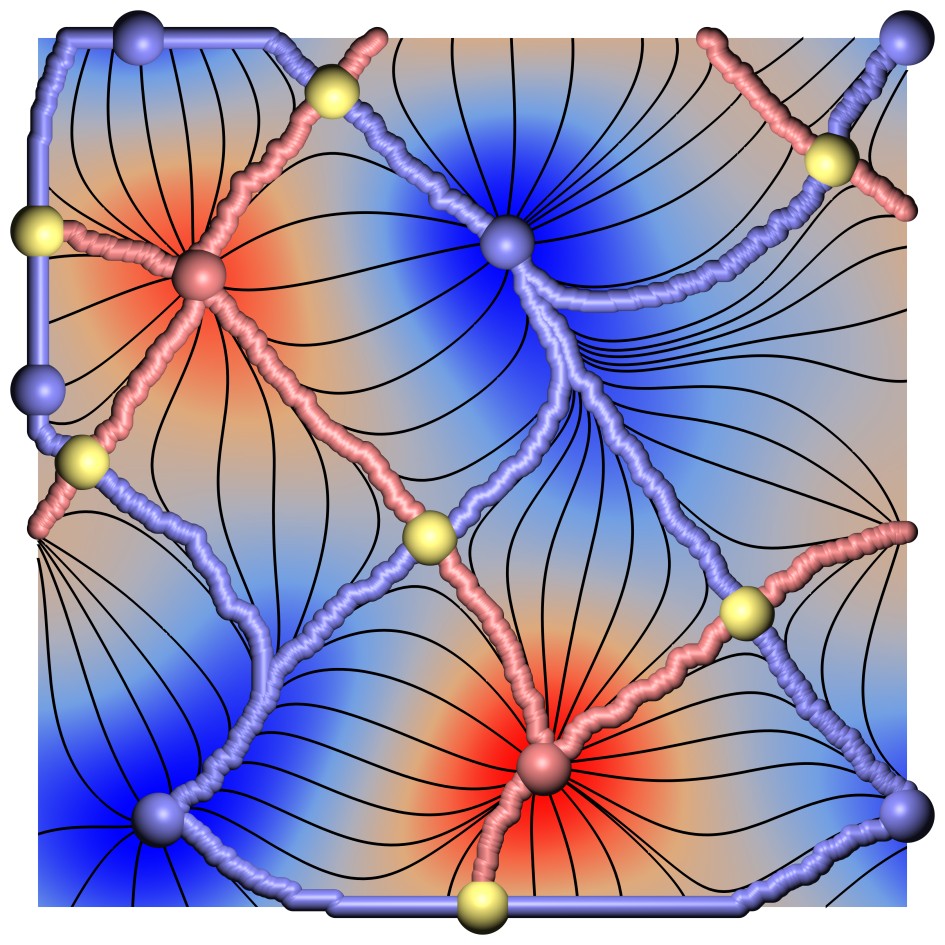}};\quad%
\node[inner sep=0pt,above right] (pic3) at (-0.49,0) {\includegraphics[width=0.24\linewidth]{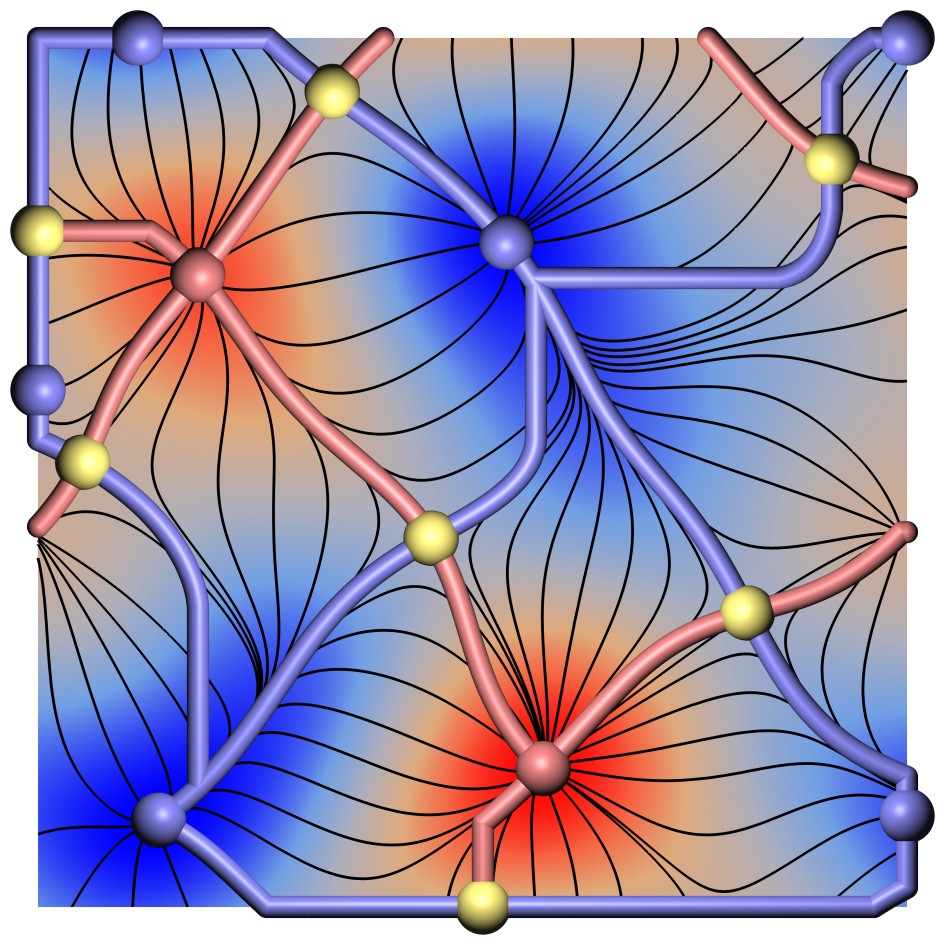}};\;%
\node[inner sep=0pt,above right] (pic4) at (-0.24,0) {\includegraphics[width=0.24\linewidth]{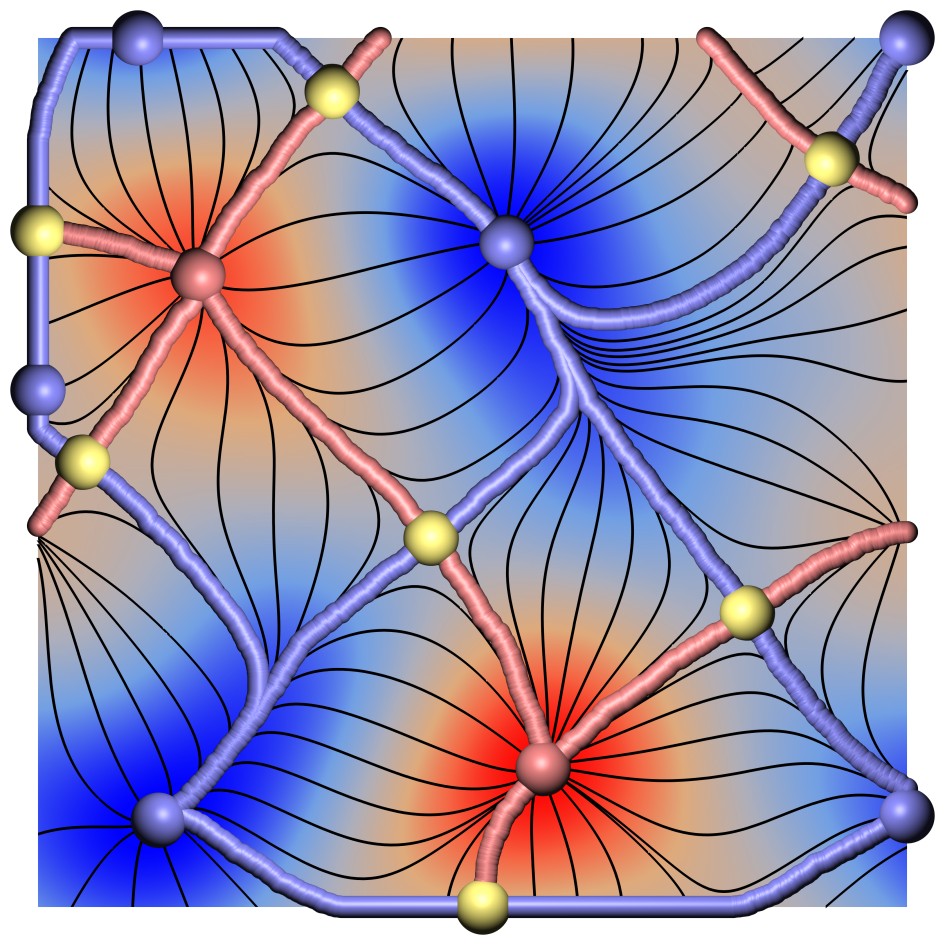}};%
\pgfsetlinewidth{0.5ex}
\pgfsetcolor{white}
\draw[-triangle 45, thick] (-0.87,0.11) -- (-0.92,0.09);
\draw[-triangle 45, thick] (-0.62,0.11) -- (-0.67,0.09);
\draw[-triangle 45, thick] (-0.37,0.11) -- (-0.42,0.09);
\draw[-triangle 45, thick] (-0.12,0.11) -- (-0.17,0.09);
\draw[-triangle 45, thick] (-0.79,0.11) -- (-0.81,0.15);
\draw[-triangle 45, thick] (-0.54,0.11) -- (-0.56,0.15);
\draw[-triangle 45, thick] (-0.29,0.11) -- (-0.31,0.15);
\draw[-triangle 45, thick] (-0.04,0.11) -- (-0.06,0.15);
\end{tikzpicture}
\\%
\begin{tikzpicture}[>=latex',join=bevel,]%
\node (text1) at (-1,0.03) {\begin{rotate}{90}Test function $2$\end{rotate}};
\node[inner sep=0pt,above right] (pic1) at (-0.99,0) {\includegraphics[width=0.24\linewidth]{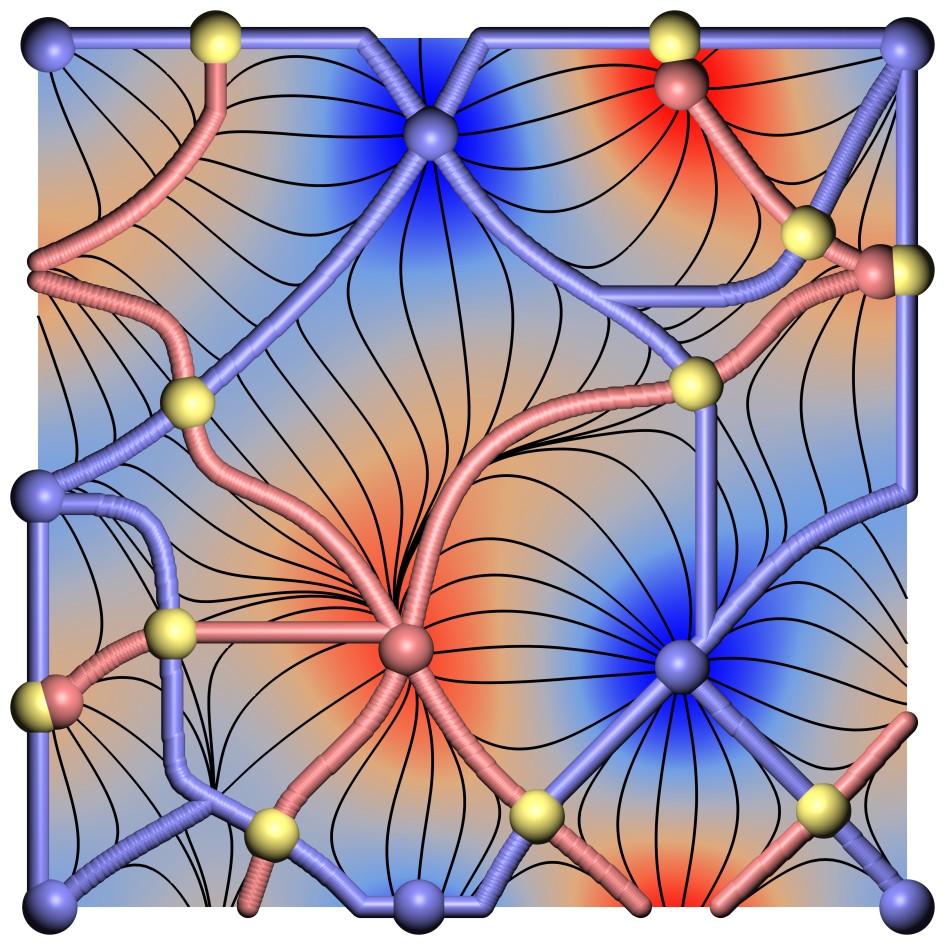}};\;%
\node[inner sep=0pt,above right] (pic2) at (-0.74,0) {\includegraphics[width=0.24\linewidth]{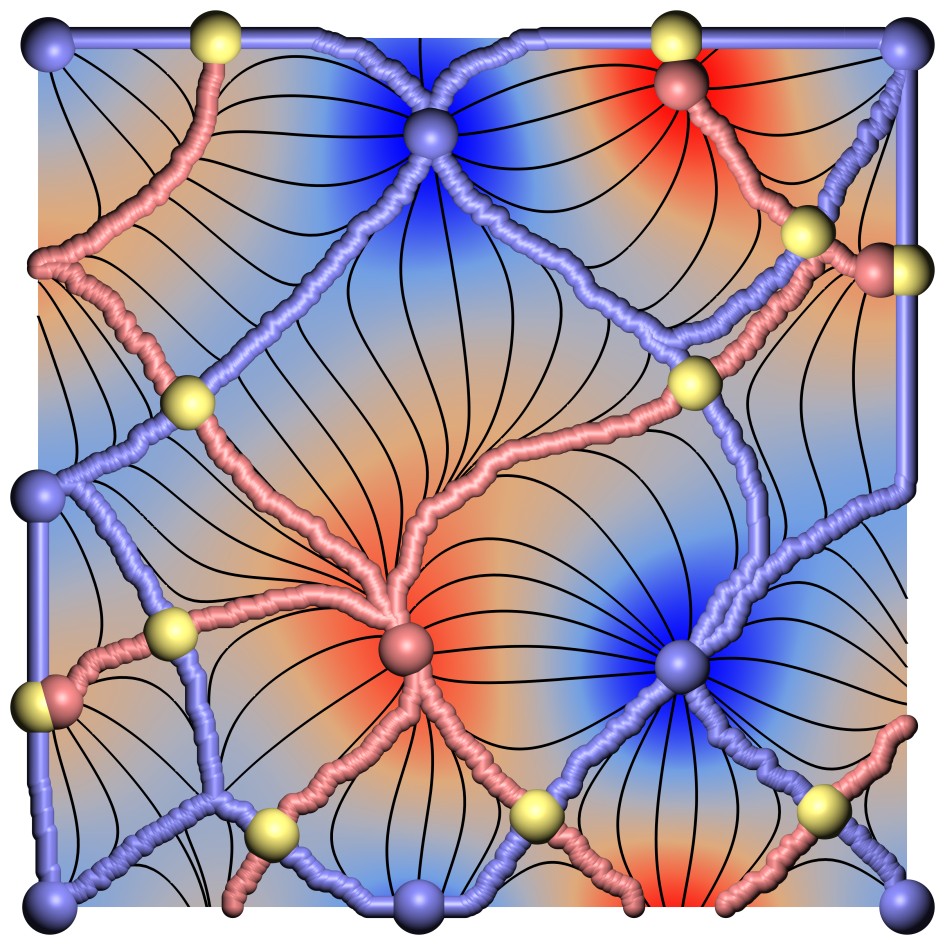}};\quad%
\node[inner sep=0pt,above right] (pic3) at (-0.49,0) {\includegraphics[width=0.24\linewidth]{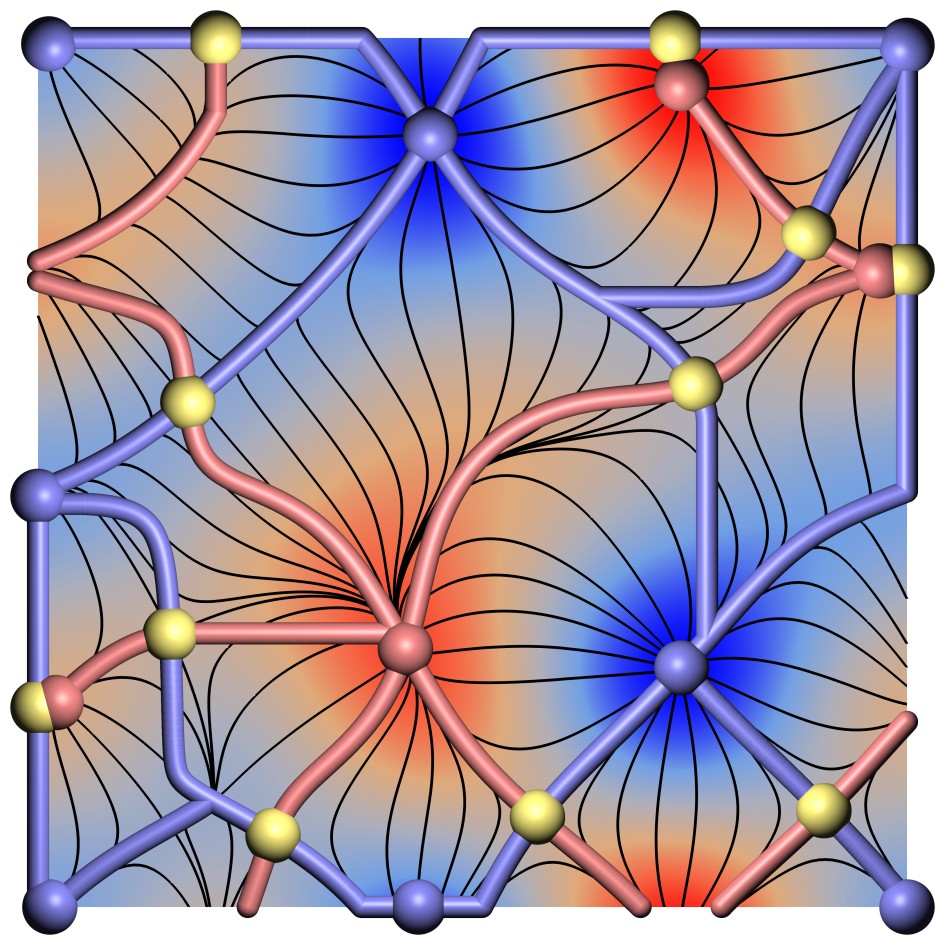}};\;%
\node[inner sep=0pt,above right] (pic4) at (-0.24,0) {\includegraphics[width=0.24\linewidth]{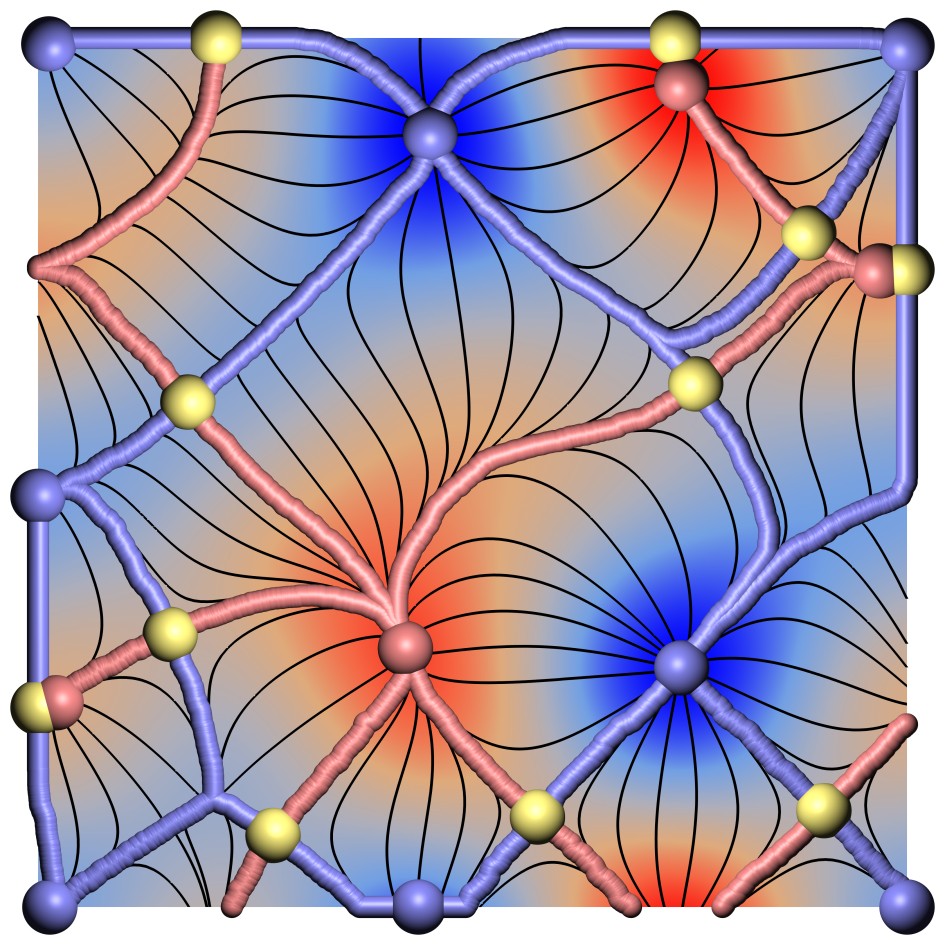}};%
\pgfsetlinewidth{0.5ex}
\pgfsetcolor{white}
\draw[-triangle 45, thick] (-0.88,0.07) -- (-0.93,0.05);
\draw[-triangle 45, thick] (-0.63,0.07) -- (-0.68,0.05);
\draw[-triangle 45, thick] (-0.38,0.07) -- (-0.43,0.05);
\draw[-triangle 45, thick] (-0.13,0.07) -- (-0.18,0.05);
\draw[-triangle 45, thick] (-0.83,0.08) -- (-0.85,0.12);
\draw[-triangle 45, thick] (-0.58,0.08) -- (-0.6,0.12);
\draw[-triangle 45, thick] (-0.33,0.08) -- (-0.35,0.12);
\draw[-triangle 45, thick] (-0.08,0.08) -- (-0.10,0.12);
\end{tikzpicture}

%\\%
%
\begin{tikzpicture}[>=latex',join=bevel,]%
\node (text1) at (-1,0.03) {\begin{rotate}{90}Test function $3$\end{rotate}};
\node[inner sep=0pt,above right] (pic1) at (-0.99,0) {\includegraphics[width=0.24\linewidth]{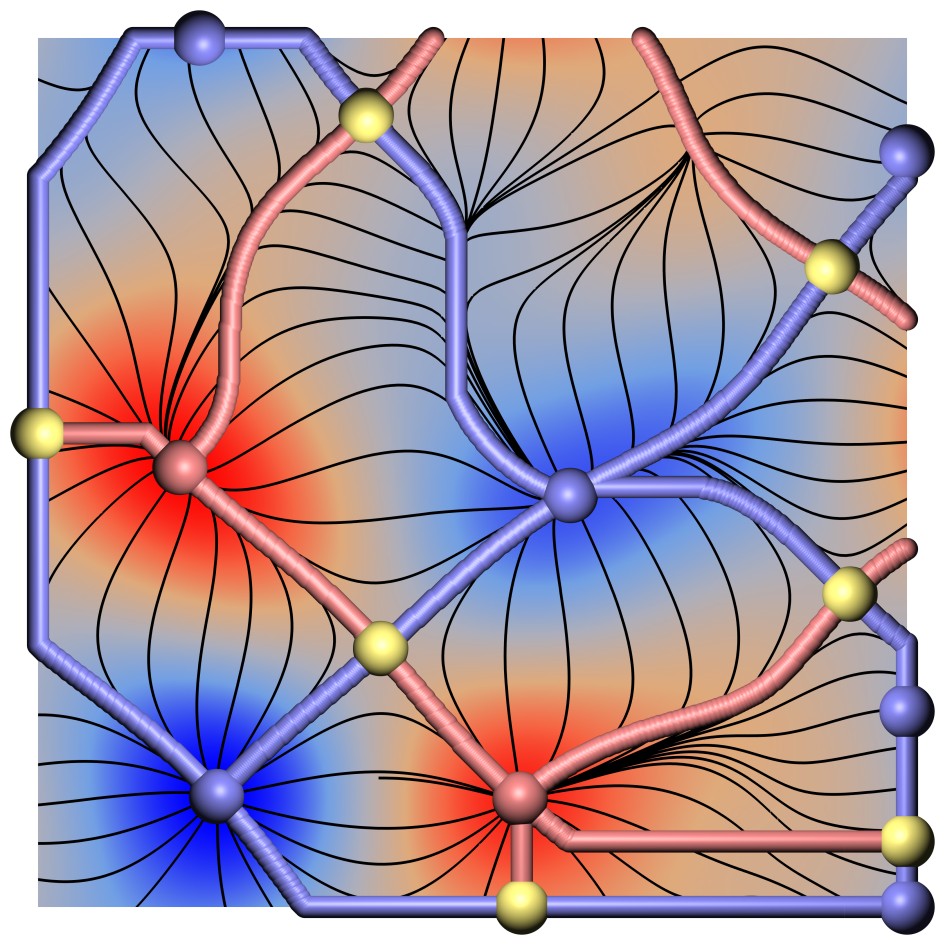}};\;%
\node[inner sep=0pt,above right] (pic2) at (-0.74,0) {\includegraphics[width=0.24\linewidth]{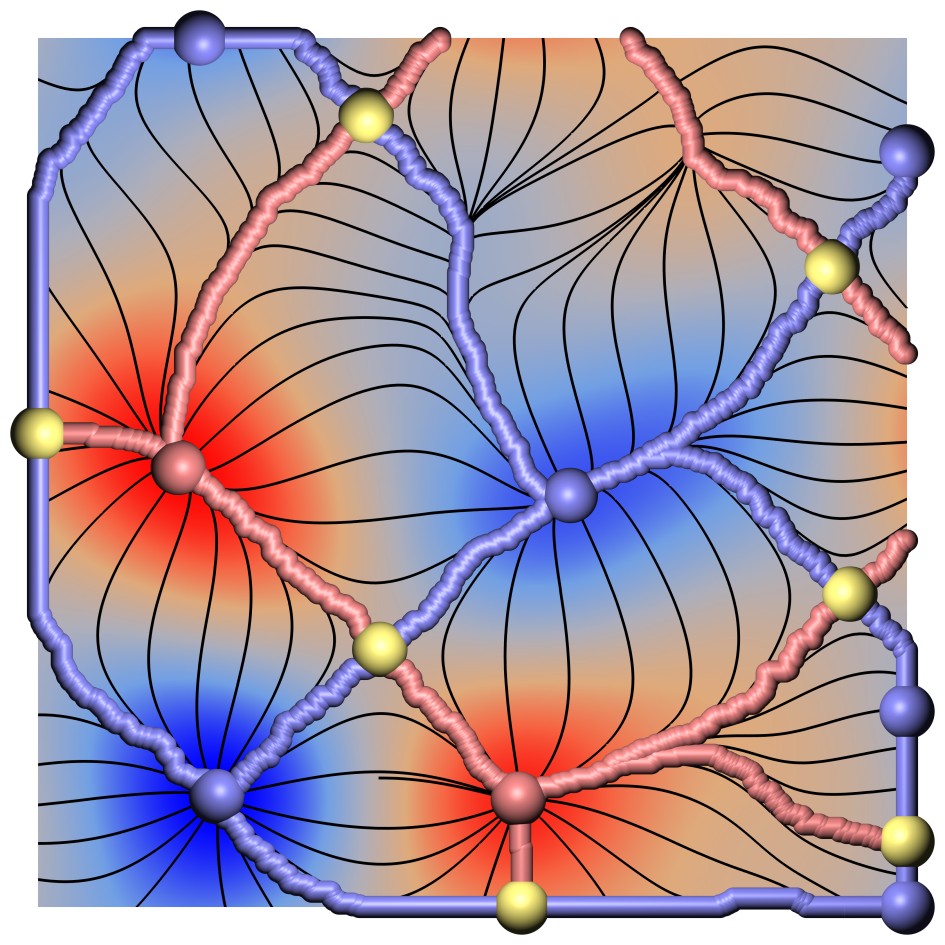}};\quad%
\node[inner sep=0pt,above right] (pic3) at (-0.49,0) {\includegraphics[width=0.24\linewidth]{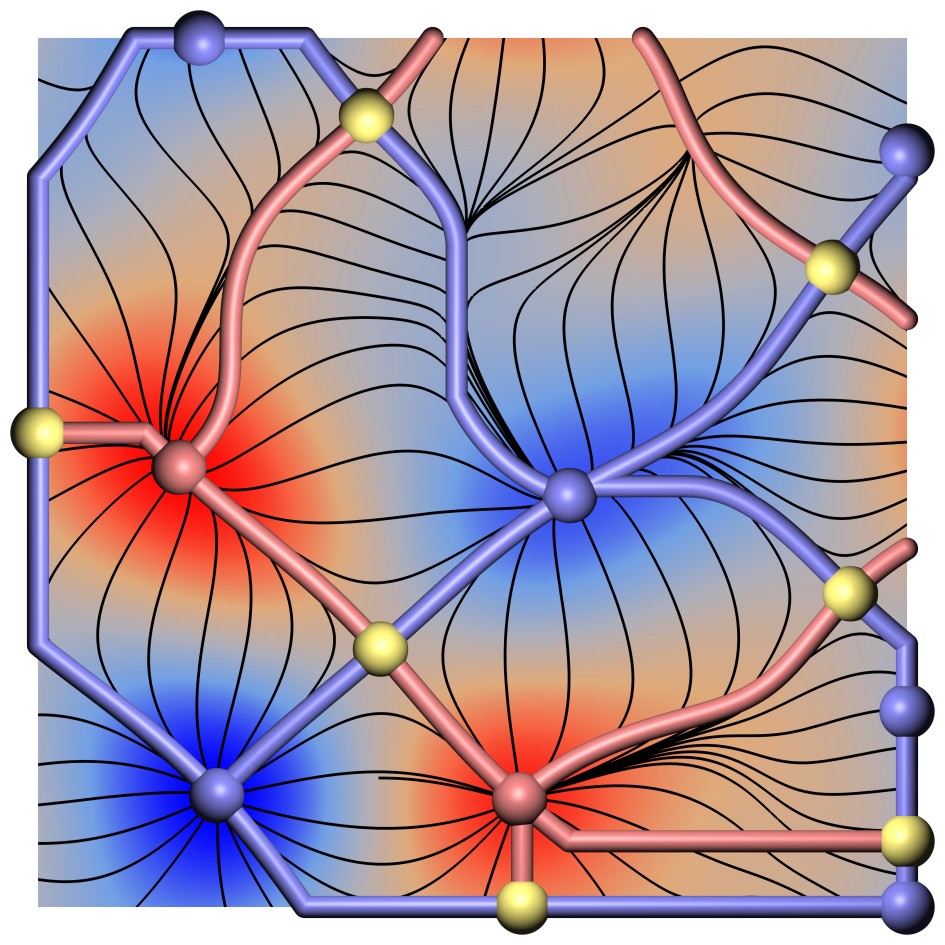}};\;%
\node[inner sep=0pt,above right] (pic4) at (-0.24,0) {\includegraphics[width=0.24\linewidth]{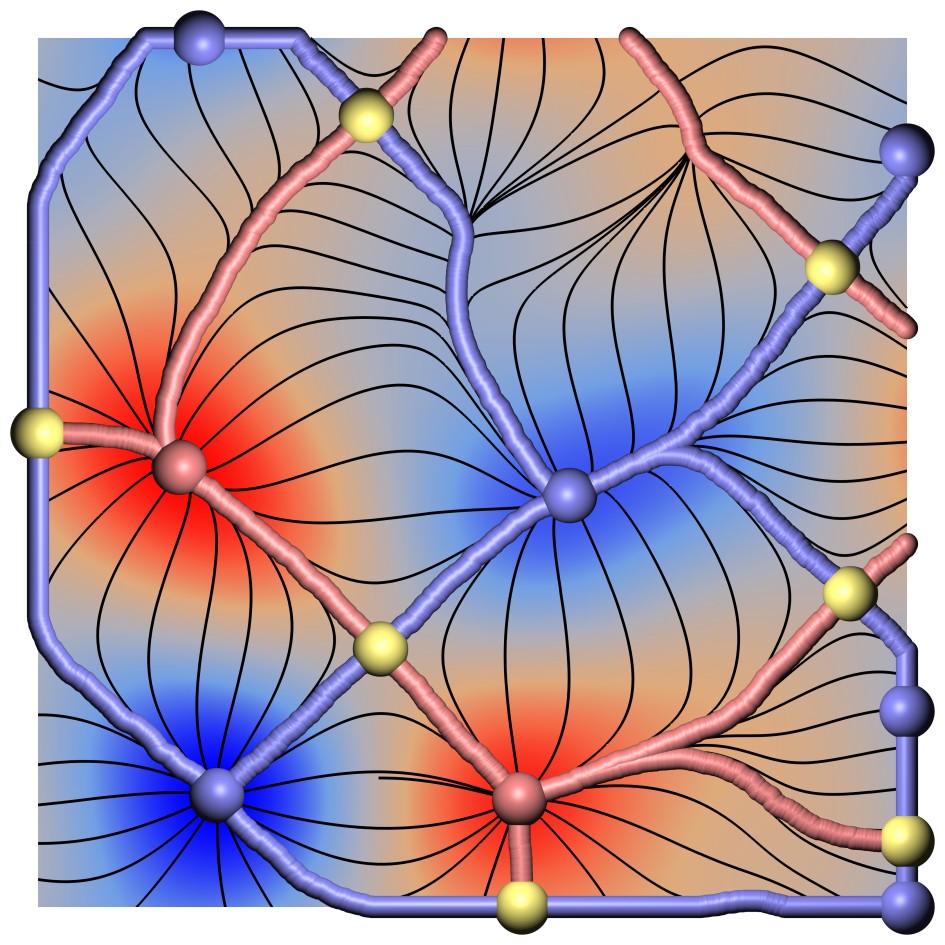}};%
\pgfsetlinewidth{0.5ex}
\pgfsetcolor{white}
\draw[-triangle 45, thick] (-0.85,0.10) -- (-0.83,0.06);
\draw[-triangle 45, thick] (-0.6,0.10) -- (-0.58,0.06);
\draw[-triangle 45, thick] (-0.35,0.10) -- (-0.33,0.06);
\draw[-triangle 45, thick] (-0.10,0.10) -- (-0.08,0.06);
\draw[-triangle 45, thick] (-0.885,0.17) -- (-0.925,0.15);
\draw[-triangle 45, thick] (-0.635,0.17) -- (-0.675,0.15);
\draw[-triangle 45, thick] (-0.385,0.17) -- (-0.425,0.15);
\draw[-triangle 45, thick] (-0.135,0.17) -- (-0.175,0.15);
\end{tikzpicture}
\\%
\begin{tikzpicture}[>=latex',join=bevel,]%
\node (text1) at (-1,0.03) {\begin{rotate}{90}Test function $4$\end{rotate}};
\node[inner sep=0pt,above right] (pic1) at (-0.99,0) {\includegraphics[width=0.24\linewidth]{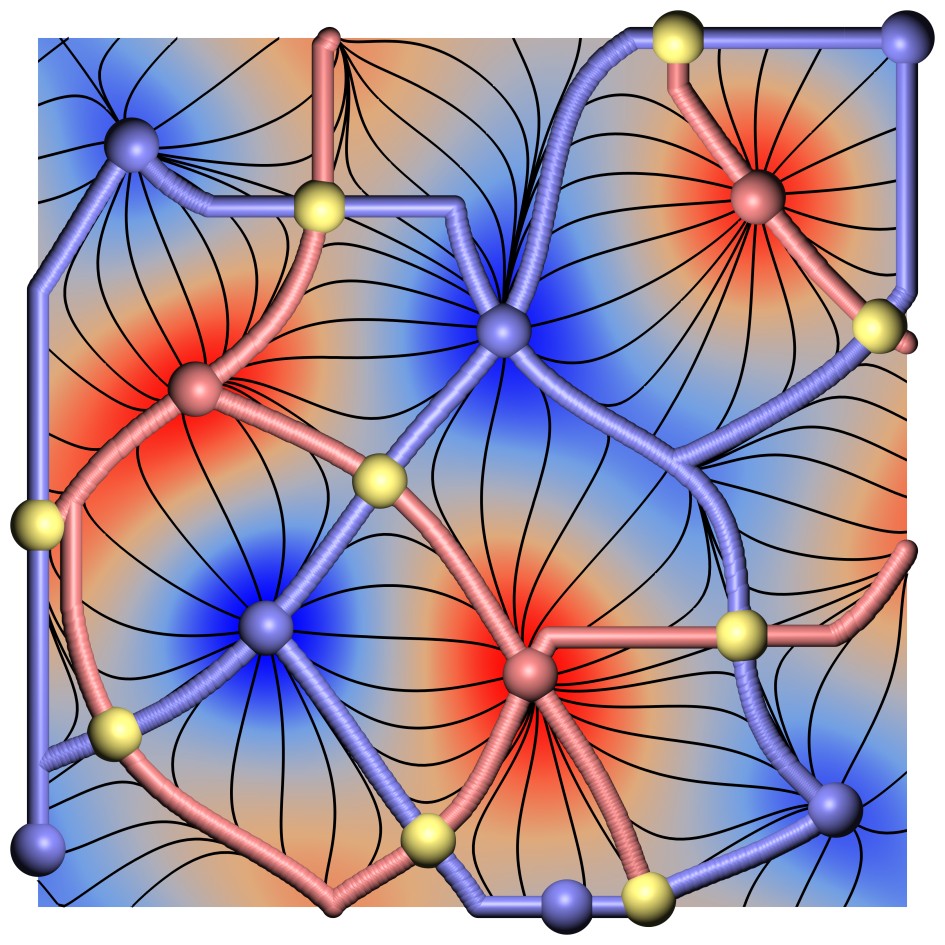}};\;%
\node[inner sep=0pt,above right] (pic2) at (-0.74,0) {\includegraphics[width=0.24\linewidth]{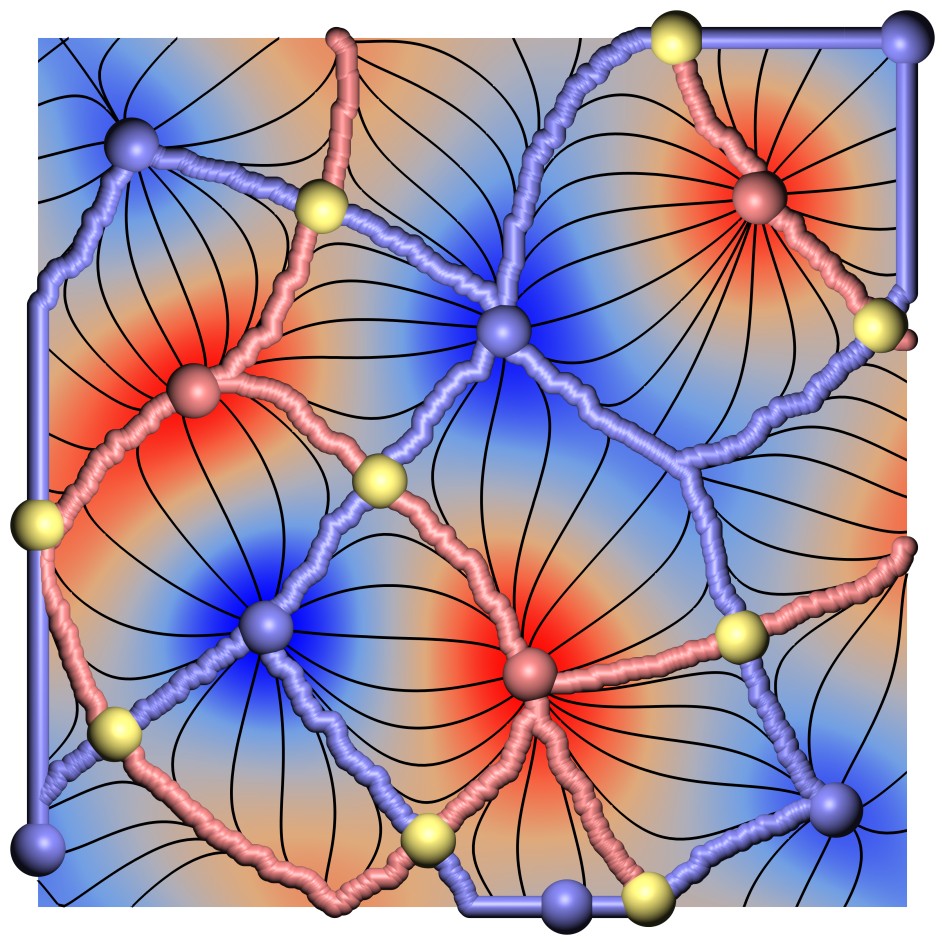}};\quad%
\node[inner sep=0pt,above right] (pic3) at (-0.49,0) {\includegraphics[width=0.24\linewidth]{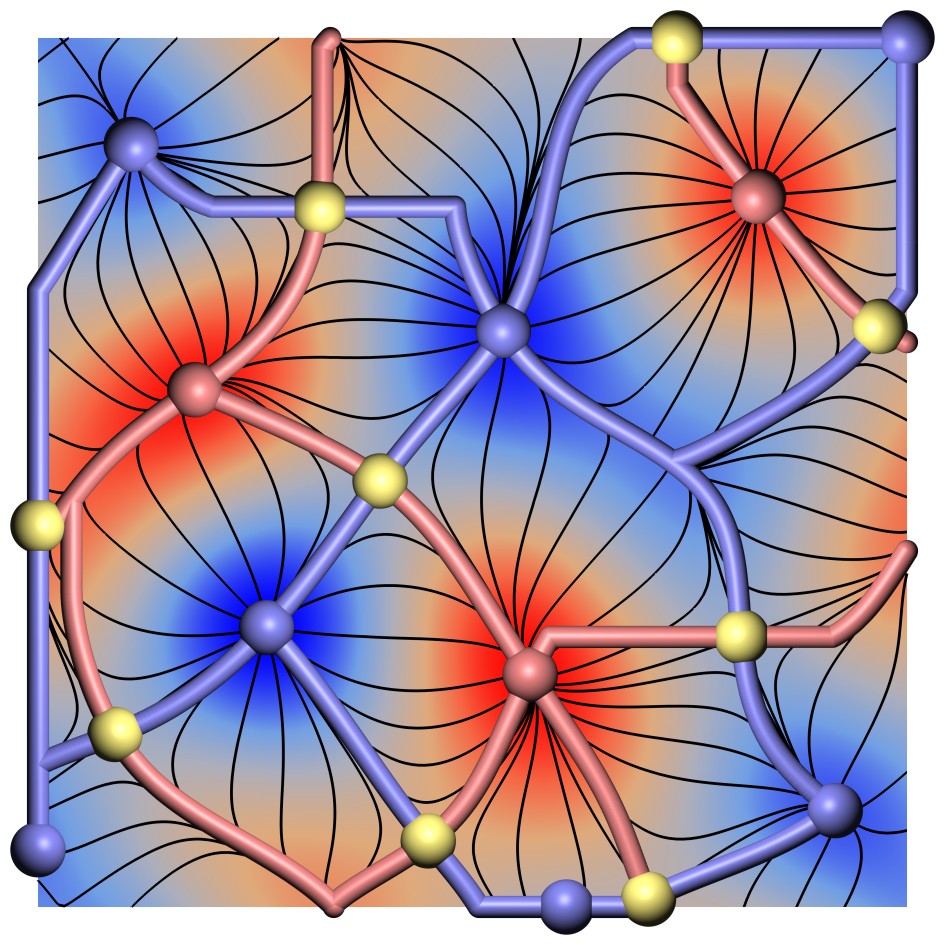}};\;%
\node[inner sep=0pt,above right] (pic4) at (-0.24,0) {\includegraphics[width=0.24\linewidth]{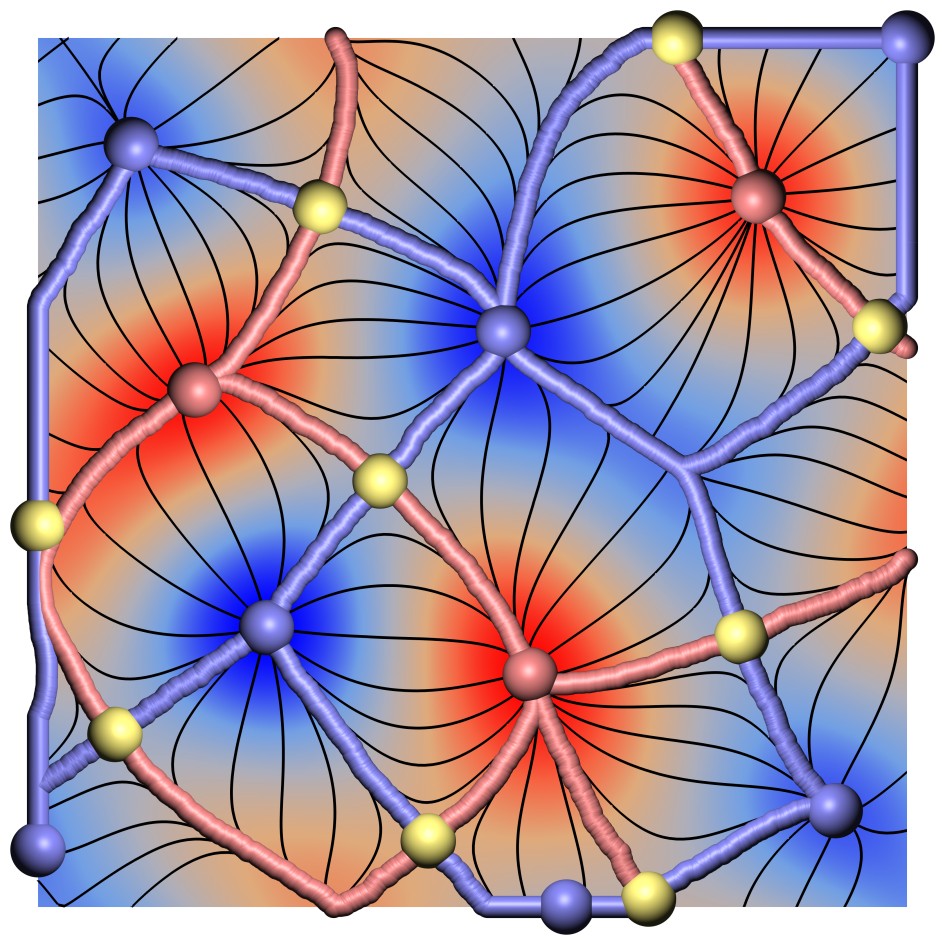}};%
\pgfsetlinewidth{0.5ex}
\pgfsetcolor{white}
\draw[-triangle 45, thick] (-0.85,0.125) -- (-0.83,0.085);
\draw[-triangle 45, thick] (-0.6,0.125) -- (-0.58,0.085);
\draw[-triangle 45, thick] (-0.35,0.125) -- (-0.33,0.085);
\draw[-triangle 45, thick] (-0.1,0.125) -- (-0.08,0.085);
\draw[-triangle 45, thick] (-0.885,0.13) -- (-0.88,0.17);
\draw[-triangle 45, thick] (-0.635,0.13) -- (-0.63,0.17);
\draw[-triangle 45, thick] (-0.385,0.13) -- (-0.38,0.17);
\draw[-triangle 45, thick] (-0.135,0.13) -- (-0.13,0.17);
\end{tikzpicture}
\caption{Comparison of the steepest descent and the probabilistic approach using generic functions. The two rows show two different functions randomly generated using expression \eqref{eq:randomField}. Red denotes a high function value, while blue denotes a low value. Black lines depict integral lines of their gradient. The left two columns show the extracted MS-complexes for each function sampled on a $512^2$ grid using the steepest descent and our probabilistic edge selection strategy, respectively. The right two columns show the result on a $4096^2$ grid. Minima, saddles and maxima are shown as blue, yellow and red spheres, whereas the 0- and 1-separatrices are shown as blue and red lines, respectively. Since separatrices are integral lines of the gradient, the blue and red lines should follow the black lines. White arrows indicate regions where the difference of the two approaches is visually apparent.}%
\label{fig:randomFields}%
\end{figure*}

\paragraph{Generic functions.}
To demonstrate that our approach is convergent for general smooth functions, we considered a set of smooth functions generated by the expression

\begin{equation}
\sum_{m,n = 1}^2 \left(X_{m,n}^{(1)}\sin(m x) + X_{m,n}^{(2)}\cos(m x)\right) \left(X_{m,n}^{(3)}\sin( n y) + X_{m,n}^{(4)}\cos( n y ) \right),
\label{eq:randomField}
\end{equation}

where the $X_{m,n}^{(j)}$'s are random variables uniformly distributed in $[-1,1]$. This expression is now evaluated on the domain $[-\pi,\pi]^2$ discretized using two different grid resolutions. We selected two representatives of this set of functions and applied the steepest descent and our probabilistic version of \alg{alg:CombinatorialGradient}. The resulting MS-complexes are shown in \fig{fig:randomFields}. 

It is apparent that the steepest descent approach does not yield the correct MS-complex. Our probabilistic edge selection strategy, however, visually converges to the correct solution. Note that we applied our method to a much larger number of such functions. We observed the above behavior in every case. 

\section{Conclusion and Future Work}
\label{sec:conclusion}

We presented a probabilistic approach that computes a combinatorial gradient for a given two dimensional image data set. Since our method is a modification of a previously proposed method\c{Robins2011}, it shares its main properties. It has a linear running time and can be applied efficiently in a parallel setting. Furthermore, its critical points correspond one-to-one with the topological changes of the sublevel sets of the input data.

Our probabilistic extension of this algorithm yields combinatorial gradients whose separatrices converge to their continuous counterpart when the grid resolution is increased. We also demonstrated that our method works for anisotropic grids and does not exhibit a grid dependency. While we only hinted at a formal proof of this property, we provided a thorough numerical evaluation and compared it to existing algorithms.

We believe that this approach could be extended and improved in several directions:
\begin{itemize}
	\item It seems that an generalization to scalar data defined on triangulated surfaces is feasible. The main challenge is thereby the generalization of the derivation of the edge selection strategy in \sec{sec:method_method}. This may also yield some insight on the representation of the discrete metric provided by the triangulation.
	\item Extending our approach to higher dimensional image data could also be interesting and useful. This may be quite challenging since combinatorial separatrices that connect the saddle points can merge and split in 3D. This stems from a fundamentally different structure of the cell graph.
	\item The efficiency of the method may be increased by developing an adaptive grid refinement approach. This may be particularly effective since a high resolution is only needed in the vicinity of the separatrices.
	\item It could also be possible to extend this idea to the combinatorial vector field context\c{Forman1998,Reininghaus_FastCVT}.
\end{itemize}

\bibliographystyle{amsalpha}

\bibliography{acsc,JanReininghaus}

\end{document}